\pdfoutput=1 
\documentclass[pdflatex,sn-mathphys]{sn-jnl}

\usepackage{amsfonts,amssymb,float}

\jyear{2021}%

\theoremstyle{thmstyleone}%
\newtheorem{theorem}{Theorem}
\theoremstyle{thmstyletwo}%

\theoremstyle{thmstylethree}%

\begin{document}

\title[A classification performance evaluation measure 
considering data separability]{A classification performance evaluation measure 
considering data separability
}


\author[1]{\fnm{Lingyan} \sur{Xue}}\email{xly1524@nudt.edu.cn}
\author*[1]{\fnm{Xinyu} \sur{Zhang}}\email{zhangxinyu90111@163.com}
\author[1]{\fnm{Weidong} \sur{Jiang}}\email{jwd2232@vip.163.com}
\author[1]{\fnm{Kai} \sur{Huo}}\email{huokai2001@163.com}

\affil[1]{\orgdiv{College of Electronic Science and Technology}, \orgname{National University of Defense Technology}, \orgaddress{\street{Deya District}, \city{Changsha}, \postcode{410000}, \state{Hunan}, \country{China}}}


\abstract{Machine learning and deep learning classification models are data-driven, and the model and the data jointly determine their classification performance. It is biased to evaluate the model's performance only based on the classifier accuracy while ignoring the data separability. Sometimes, the model exhibits excellent accuracy, which might be attributed to its testing on highly separable data. Most of the current studies on data separability measures are defined based on the distance between sample points, but this has been demonstrated to fail in several circumstances. In this paper, we propose a new separability measure—the rate of separability (RS), which is based on the data coding rate. We validate its effectiveness as a supplement to the separability measure by comparing it to four other distance-based measures on synthetic datasets. Then, we  demonstrate the positive correlation between the proposed measure and recognition accuracy  in a multi-task scenario constructed from a real dataset. Finally, we  discuss the methods for evaluating the classification performance of machine learning and deep learning models considering data separability.}




\maketitle

\section{Introduction}\label{sec1}

As an important branch in data mining, classification aims to construct a classification model to learn a mapping regularity from existing data to class labels. The research of model is essential, yet data also determines the performance \cite{bib1}. A specific example is the impact of spectral separability on classification accuracy \cite{bib2}.Numerous classification models have been proposed, including KNN, SVM, logistic regression, neural networks, etc., but studies on data separability are substantially fewer. A recent study in hyperspectral image classification has argued that insufficient data may limit the assessment capability of existing accuracy indexes \cite{bib3}. That leads to the problem of whether a model performs best on a classification case is unclear or inconclusive \cite{bib4,bib5}. It is acknowledged that a good classification model provides greater generalization potential, which means finding rules consistent with available data that apply widely to predict the class of unknown data \cite{bib6}. Yet the criteria for assessing the model’s generalization ability remain debated. To simplify the performance evaluation process, researchers generally tend to adopt measures based on the confusion matrix \cite{bib7}, like accuracy, precision, kappa statistic, and F-score. Each measure is represented with a single score number, making it straightforward to compare and analyze classification models quantitatively. Although the result is intuitive, its comparability is invalid when confronted with a multi-task classification situation more representative of the real-world environment.

A contradictory example is that a classifier reaches the highest accuracy in one task but the lowest in another. What causes the problem is that such classifier-oriented measures treat the different instances of a dataset as statistical objects and ignore the classification difficulty of each instance. For the above issue, Yu et al. \cite{bib8} proposed an instance-oriented measure but only apply to data with few samples due to the computational complexity of classification difficulty for each instance. Therefore, we require a measure to statistically characterize the classification difficulty of datasets. Fortunately, previous research has established that separability is an intrinsic characteristic of a dataset \cite{bib9} to describe how instances belonging to different classes mix.Measuring the data quality is critical for estimating the problem’s difficulty in advance since a classification model’s accuracy strongly depends on the data quality \cite{bib10}. Obviously, the more separable the dataset, the simpler the classification. Eventually, we consider data separability as a metric of classification difficulty. 

Such a metric would be helpful in the following ways: First, the metric is used to evaluate the dataset. Specifically, experts can use it to build a new standard dataset according to a list of desired levels of classification difficulty. In addition, if necessary, the classification difficulty can be adjusted by modifying the dataset. Second, this metric aids in selecting the dataset most appropriate for the system objectives. And even if only one dataset is available, it is beneficial to know its classification difficulty in advance. Researchers can predict the range of classification accuracy based on difficulty information. If the precision is lower than the expected value, the algorithm's utilization may be wrong. Finally, the performance evaluation can be more objective. For instance, if the metric of a personal dataset and its performance results are provided simultaneously, we can fairly judge the model's performance by comparing it to that of a standard dataset with known classification difficulty. Similarly, the performance on two recognition tasks with varying classification difficulty will be comparable.

There are several measures of data separability that can quantify classification difficulty. Fisher discriminant ratio (FDR) \cite{bib11} has been used in many studies, which measures the data separability using each class's mean and standard deviation. Similarly inspired by FDR, Generalized Discrimination Value (GDV) \cite{bib12}, which has dimensional invariance, is also proposed to quantify the separability of data classes in neural networks. But the above feature-based measures are proved to fail in some cases, like two-class circle data in Figure \ref{fig3}(c).

A more practical issue is data complexity, which primarily measures the Euclidean distance between intra-class and inter-class data. Ho and Basu \cite{bib13} conducted a groundbreaking review of data complexity measures. Recently, Lorena et al. \cite{bib14} reviewed existing methods for measuring classification complexity, showing that some of those may have enormous time costs. The measures above work based on the linearity assumption of data and, therefore, do not identify properly clusters characterized by nonlinear structures. To overcome this limitation, Maria and Paolo \cite{bib15} summarized several approaches: density-, kernel-, graph- or manifold-based methods. However, such distance-based methods for assessing data separability still leave an open question, namely, in high dimensional space, the boundary conditions of the data to which various distance metrics (such as Euclidean, Manhattan, and Minkowski distance) are applicable \cite{bib16}.

As an alternative to the distance-based criterion, we consider explaining data separability from the perspective of probability theory. According to the Bayesian criterion, when each category has an equal probability of occurrence, the result of sample partitioning depends on the magnitude of the likelihood function corresponding to each class. At this point, the more significant the gap between the likelihood functions of the various categories, the more separable the data. However, due to the complexity of high-dimensional data, the form of the likelihood function and parameter estimation are uncertain, which brings low classification accuracy. As inspired by Cover and Thomas  \cite{bib17}, the process of minimizing the data rate distortion is equivalent to the operation of solving the optimal solution of the likelihood function, i.e., the data coding rate has strong consistency with the parameter estimation performance  \cite{bib18}. That means if the data can be fitted with a better distribution model after segmentation, then the data should be effectively encoded with such a model. Ma et al. \cite{bib19} argued that the coding rate (subject to distortion) provides a natural measure of the goodness of segmentation for real-valued mixed data.

Since there is no research verifying the feasibility of using coding rate as a measure of data separability, this is the first study to construct a separability measure based on rate-distortion theory called the rate of separability (RS). The main contributions of this paper are summarized as follows.

\begin{itemize}
	\item[$\bullet$] We proposed a data separability measure based on rate-distortion theory, derived a new form applied to non-zero mean data, and verified its effectiveness in theory and experiments.
\end{itemize}

\begin{itemize}
	\item[$\bullet$] We found a positive correlation between classification accuracy and data separability in a multi-task noisy environment. 
\end{itemize}

\begin{itemize}
	\item[$\bullet$] In a multi-task noisy environment, we designed a task-oriented classifier performance evaluation method considering data separability as the task difficulty. Unlike the classification accuracy changing with different tasks, this method obtains classifier ability as the classifier’s inherent property under certain assumptions. 
\end{itemize}

\begin{itemize}
	\item[$\bullet$] We built a modular classifier performance evaluation model to explain the function of deep learning convolutional blocks using data separability.
\end{itemize}

The rest of this paper is structured as follows. Section \ref{sec2} introduces the method of constructing coding-rate-based measure and the theory of distance-based measures. Section \ref{sec3} provides experimental methods for validating measure validity. Section \ref{sec4} evaluates the classification model performance; results and analysis are also given in this section. Finally, we conclude in Section \ref{sec5}.

\section{Measures of data separability}\label{sec2}

Whether or not data is challenging to separate depends on its confusion level. For example, if data points from the interclass are mixed, and data points from the intra-class are scattered, the data is difficult to classify. Such a condition appears to be described by the inter and intra-class distance. Specifically, we expect the intraclass data to be closer and the inter-class data to be further away. Guan et al. \cite{bib20} recently demonstrated that several data separability measures (e.g., DSI, N2, LSC, Density) based on the inter and intra-class distance perform well in a two-class dataset. Due to the calculation of the inter and intra-class distance, these algorithms may have to traverse almost every data point at times, resulting in massive computation. 

Another point of view comes from the information theory. A system's entropy can be used to measure how chaotic it is. Similarly, data's level of confusion can be defined by data's entropy. When the data points are scattered in various regions of the space, the entropy of the data can be calculated by the proportion of the data occupying the region \cite{bib21}. Unfortunately, entropy is not suitable for continuous random variables or high-dimensional space. Another information theory concept that quantifies the "compactness" of a random distribution was developed to address this problem. It is known as rate-distortion \cite{bib17}. Notably, it can be accurately and efficiently computed in closed form. Ma et al. \cite{bib19} provide a precise estimation of rate-distortion on a subspace and a mixture of subspaces. In recent studies, rate-distortion has been used to explain or construct deep networks \cite{bib22, bib23}. Yet there is no research verifying the feasibility of using rate-distortion as a measure of data separability. 

Therefore, part of the work of this paper is to verify the effectiveness of the proposed separability measure based on rate-distortion theory and compare it with distance-based measures. In this section, we briefly introduce the distance-based separability measures and then apply rate-distortion theory in constructing a new data separability measure.

\subsection{Distance based data separability measures}\label{subsec2.1}

Here we introduce several distance-based data separability measures which effectively assess typical two-class datasets \cite{bib1}.

\textbf{1). DSI}

DSI was built by measuring the Kolmogorov-Smirnov similarity between intra-class distance distribution $\{ {d_{{{\bf{X}}^j}}}\} $  and inter-class distance distribution $\{ {d_{{{\bf{X}}^j},{{{\bf{\bar X}}}^j}}}\} $ :

\begin{equation}
DSI = \frac{1}{k}\sum\limits_{j = 1}^k {KS(\{ {d_{{{\bf{X}}^j}}}\} ,\{ {d_{{{\bf{X}}^j},{{{\bf{\bar X}}}^j}}}\} )}.\label{eq1}
\end{equation}

\textbf{2). N2}

To calculate N2, we first compute the Euclidean distance between each point ${{\bf{x}}_i}$ , its nearest neighbor  $NN({{\bf{x}}_i})$ and its nearest enemy $NE({{\bf{x}}_i})$  . Then we have  $d({{\bf{x}}_i},NN({{\bf{x}}_i}))$ and $d({{\bf{x}}_i},NE({{\bf{x}}_i}))$ , take the sum of all the distances to $d({{\bf{x}}_i},NN({{\bf{x}}_i}))$  and $d({{\bf{x}}_i},NE({{\bf{x}}_i}))$ , the ratio of two sum results is defined as N2:

\begin{equation}
N2 = \frac{{\sum\limits_{i = 1}^m {d({{\bf{x}}_i},NN({{\bf{x}}_i}))} }}{{\sum\limits_{i = 1}^m {d({{\bf{x}}_i},NE({{\bf{x}}_i}))} }}.\label{eq2}
\end{equation}

\textbf{3). LSC}

The Local Set (LS) includes the points ${{\bf{x}}_j}(j \ne i)$  whose distance to ${{\bf{x}}_i}$  is smaller than the distance from ${{\bf{x}}_i}$  to  ${{\bf{x}}_i}$’s nearest enemy $NE({{\bf{x}}_i})$ . 

\begin{equation}
LS({{\bf{x}}_i}) = \{{{\bf{x}}_j}\vert d({{\bf{x}}_i},{{\bf{x}}_j}) < d({{\bf{x}}_i},NE({{\bf{x}}_i}))\}\label{eq3}
\end{equation}

This measure is to count the number of ${{\bf{x}}_i}$  . 

\begin{equation}
LSC = 1 - \frac{1}{{{m^2}}}\sum\limits_{i = 1}^m {\vert LS({{\bf{x}}_i})\vert}\label{eq4}
\end{equation}

\textbf{4). Density}

To use this measure, it is necessary to represent the classification dataset as a graph $G = (V,E)$  where $\vert V \vert = m$ , $0 \le \vert {\rm E} \vert \le \frac{{m(m - 1)}}{2}$ . Two nodes  ${{\bf{x}}_i}$ and ${{\bf{x}}_j}$  are connected only if $d({{\bf{x}}_i},{{\bf{x}}_j}) < \varepsilon$ , then the measure is given by:

\begin{equation}
Density = 1 - \frac{{2 \vert E \vert}}{{m(m - 1)}}.\label{eq5}
\end{equation}

\textbf{5). transform all measures into [0,1]}

To facilitate comparability, we transform the values of some of the measures so that all range in [0, 1] with low values meaning high separability. Since RS, LSC, and Density have fulfilled the conditions, the modifications of DSI and N2 are as follows.

The DSI values are already in the range [0, 1], but the low values indicate low separability. Hence, we modify it as
	 
\begin{equation}
DSI \leftarrow 1 - DSI.\label{eq6}
\end{equation}
originally, the N2 values are in range $[0,\infty )$. Hence, we shall use

\begin{equation}
N2 \leftarrow 1 - \frac{1}{{1 + N2}}.\label{eq7}
\end{equation}

\subsection{Coding-rate-based data separability measure}\label{subsec2.2}

\subsubsection{definition and computation of the coding rate}\label{subsubsec2.2.1}

According to Cover and Thomas’ \cite{bib17} definition of rate-distortion: given a data ${\bf{X}}=[{{\bf{x}}_1},...,{{\bf{x}}_m}] \in {\mathbb{R}^{d \times m}}$ with $m$ samples of $d$  dimension  and a encoding precision $\varepsilon  > 0$, the rate-distortion ${\rm{R}}({\bf{X}},\varepsilon )$ is the minimal number of binary bits needed to encode $\bf{X}$ and the expected decoding error is less than $\varepsilon$. The actual estimation coding rate of $\bf{X}$  with zero mean is as follows:

\begin{equation}
{\rm{R}}({\bf{X}},\varepsilon ) = \frac{m}{2}\log \det ({\bf{I}} + \frac{d}{{m{\varepsilon ^2}}}{\bf{X}}{{\bf{X}}^T}).\label{eq8}
\end{equation}

Furthermore, suppose $\bf{X}$ has  $k$-class samples, then ${\bf{X}} = {{\bf{X}}^1} \cup {{\bf{X}}^2} \cup ... \cup {{\bf{X}}^k}$ . the data ${{\bf{X}}^j}$   in each class $j$   also occupy a certain volume in its low dimensional subspace. For each subset, the above coding rate (\ref{eq8}) is applied. To determine the per-class sample's overall coding rate ${{\rm{R}}_{\rm{C}}}({\bf{X}},\varepsilon \vert {\bf{\Pi}})$ , let ${\bf{\Pi }}=\{ {{\bf{\Pi }}^j}\in{\mathbb{R}^{m \times m}}\} _{j = 1}^k$  be the label matrix of the $\bf{X}$  in the $k$  classes, and ${{\bf{\Pi }}^j}(i,i)$  is the probability of  ${{\bf{x}}_i}$ belonging to class $j$ , then ${{\rm{R}}_{\rm{C}}}({\bf{X}},\varepsilon \vert{\bf{\Pi }})$  is given by

\begin{equation}
\label{eq9}
\begin{aligned}
       \sum\limits_{j = 1}^k {{\rm{R}}({{\bf{X}}^j},\varepsilon )} &= {{\rm{R}}_C}({\bf{X}},\varepsilon \vert {\bf{\Pi }}) \\
       &= \sum\limits_{j = 1}^k {\frac{{tr({{\bf{\Pi }}^j})}}{2}} \log \det \left( {{\bf{I}} + \frac{d}{{tr({{\bf{\Pi }}^j}){\varepsilon ^2}}}{\bf{X}}{{\bf{\Pi }}^j}{{\bf{X}}^T}} \right)
\end{aligned}
\end{equation}

\begin{figure}[h]%
\centering
\includegraphics[width=0.9\textwidth]{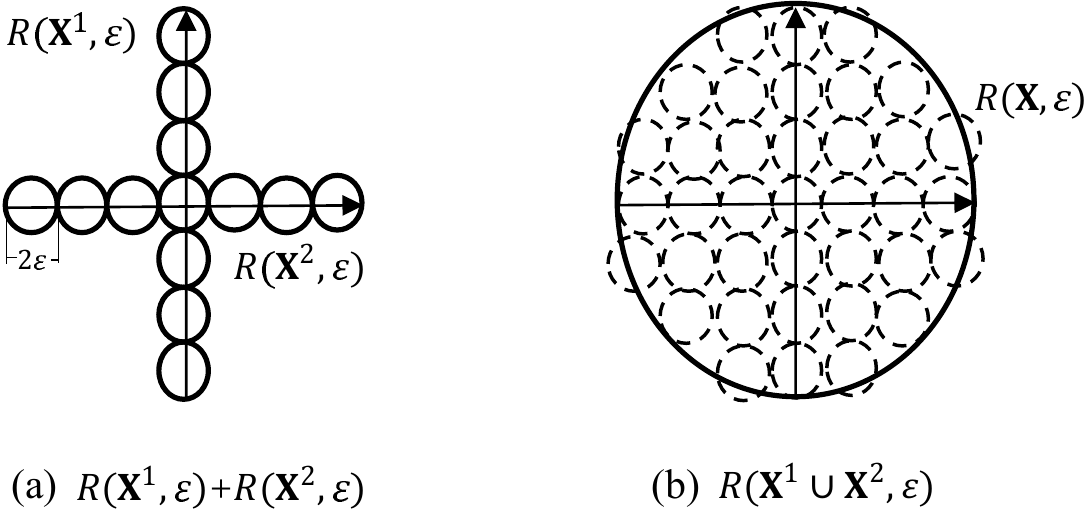}
\caption{The coding rate of $\bf{X}^j$  and $\bf{X}$ . Taking a  $\varepsilon$-ball as an encoding unit, the volume of the space spanned by $\bf{X}^j$  and $\bf{X}$   can be expressed by the number of the  $\varepsilon$-balls. The bold solid line represents the volume of both.}\label{fig1}
\end{figure}

The difference between ${\rm{R}}({{\bf{X}}^1},\varepsilon ) + {\rm{R}}({{\bf{X}}^2},\varepsilon )$ and ${\rm{R}}({{\bf{X}}^1} \cup {{\bf{X}}^2},\varepsilon )$ can be described as Figure \ref{fig1} in a two-class dataset. 

\subsubsection{Derivation of coding rate for non-zero mean data}\label{subsubsec2.2.2}

The equation for the coding rate in Section \ref{subsubsec2.2.1} is for the scenario where the mean value of the given data is zero mean. More generally, in this section, we derive the formula for the coding rate when the data's mean is non-zero.

When ${\bf{X}}=[{{\bf{x}}_1},...,{{\bf{x}}_m}] \in {\mathbb{R}^{d \times m}}$  is not zero mean, we have ${\boldsymbol {\mu }} = \frac{1}{m}\sum\limits_{i = 1}^m {{{\bf{x}}_i} \in {\mathbb{R}^d}}$ . Here we disassemble the data into two components: the zero mean part of the data and the mean of the data, that is ${\bf{X}} = {\bf{\bar X}} + {\bf{V}}$ . And we define ${\bf{V}} = {\boldsymbol {\mu }} \cdot {{\bf{1}}_{1 \times m}} = [{\bf{\mu }},{\bf{\mu }},...,{\bf{\mu }}] \in {\mathbb{R}^{d \times m}}$ . Then, the error induced on the matrix  $\bf{X}$ is

\begin{equation}
\delta {\bf{X}} = \delta {\bf{\bar X}} + \delta {\bf{V}} = \delta {\bf{\bar X}} + \delta {\boldsymbol {\mu }} \cdot {{\bf{1}}_{1 \times m}}.\label{eq10}
\end{equation}

Supposing that $\delta {\bf{\bar X}}$  and  $\delta {\bf{V}}$ are zero-mean independent random variables, the expected total squared error is

\begin{equation}
    {\rm E}\left( {{\rm{tr}}\left( {\delta {\bf{X}}\delta {{\bf{X}}^T}} \right)} \right) = {\rm E}\left( {{\rm{tr}}\left( {\delta {\bf{\bar X}}\delta {{{\bf{\bar X}}}^T}} \right)} \right) + m{\rm E}\left( {{\rm{tr}}\left( {\delta {{\boldsymbol {\mu }}^T}\delta {\boldsymbol {\mu }}} \right)} \right).\label{eq11}
\end{equation}

Now, we define the precision $\varepsilon ' = \frac{\varepsilon }{{\sqrt d }}$  encoding each entry: ${{\bar x}_{ij}}$  and ${\mu _i}$ . And the error $\delta {{\bar x}_{ij}}$ and $\delta {\mu _i}$ are uniformly distributed in the interval $[ - \frac{\varepsilon }{{\sqrt d }},\frac{\varepsilon }{{\sqrt d }}]$ . We can compute the expected total squared error of ${{\bf{\bar X}}}$  and  ${\boldsymbol {\mu }}$ are

\begin{align}
{\rm E}\left( {{\rm{tr}}\left( {\delta {\bf{\bar X}}\delta {{{\bf{\bar X}}}^T}} \right)} \right) = {\rm E}\left( {\sum\limits_{i,j} {\delta \bar x_{ij}^2} } \right) = m \times d \times \frac{{{{\left( {\frac{{2\varepsilon }}{{\sqrt d }}} \right)}^2}}}{{12}} &= \frac{{m{\varepsilon ^2}}}{3} \nonumber \\
{\rm E}\left( {{\rm{tr}}\left( {\delta {{\boldsymbol {\mu }}^T}\delta {\boldsymbol {\mu }}} \right)} \right) = {\rm E}\left( {\sum\limits_i {\delta \mu _i^2} } \right) = d \times \frac{{{{\left( {\frac{{2\varepsilon }}{{\sqrt d }}} \right)}^2}}}{{12}} = \frac{{{\varepsilon ^2}}}{3}.\label{eq12}
\end{align}

The total squared error satisfies

\begin{equation}
    {\rm E}\left( {{\rm{tr}}\left( {\delta {\bf{X}}\delta {{\bf{X}}^T}} \right)} \right) = \frac{{m{\varepsilon ^2}}}{3} + \frac{{m{\varepsilon ^2}}}{3} = \frac{{2m{\varepsilon ^2}}}{3} < m{\varepsilon ^2}.\label{eq13}
\end{equation}

Then, the mean squared error per vector in $\bf{X}$  is

\begin{equation}
    \frac{1}{m}{\rm E}\left( {{\rm{tr}}\left( {\delta {\bf{X}}\delta {{\bf{X}}^T}} \right)} \right) < {\varepsilon ^2}.\label{eq14}
\end{equation}

Since $\varepsilon ' = \frac{\varepsilon }{{\sqrt d }}$ is zero mean, we apply the conclusion in formula (\ref{eq8}), its coding rate with precision $\varepsilon ' = \frac{\varepsilon }{{\sqrt d }}$ is

\begin{equation}
    {\rm{R}}({\bf{\bar X}}) = \frac{1}{2}\log \det ({\bf{I}} + \frac{d}{{m{\varepsilon ^2}}}{\bf{\bar X}}{{{\bf{\bar X}}}^T}).\label{eq15}
\end{equation}

And the number of bits to encoding the mean vector ${\boldsymbol {\mu }}$  with precision $\varepsilon ' = \frac{\varepsilon }{{\sqrt d }}$  is

\begin{equation}
 \label{eq16}
 \begin{aligned}
        {\rm{R}}({\boldsymbol {\mu }}) &= \sum\limits_{i = 1}^d {\frac{1}{2}{{\log }_2}\left( {1 + {{\left( {\frac{{{\mu _i}}}{{\varepsilon '}}} \right)}^2}} \right)}\\
        &= \frac{1}{2}\sum\limits_{i = 1}^d {{{\log }_2}\left( {1 + \frac{{d\mu _i^2}}{{{\varepsilon ^2}}}} \right)}  \le \frac{d}{2}{\log _2}\left( {1 + \frac{{{{\boldsymbol {\mu }}^T}{\boldsymbol {\mu }}}}{{{\varepsilon ^2}}}} \right). 
\end{aligned}
\end{equation}

The above inequality is from the following formula (\ref{eq17})

\begin{equation}
\begin{aligned}
\label{eq17}
&\frac{{{{\log }_2}(1 + {\mu _1}) + {{\log }_2}(1 + {\mu _2}) +  \cdots  + {{\log }_2}(1 + {\mu _d})}}{d} \\
&\le {\log _2}\left( {1 + \frac{{{\mu _1} + {\mu _2} +  \cdots {\mu _d}}}{d}} \right),    
\end{aligned}
\end{equation}

for nonnegative real numbers ${\mu _1},{\mu _2},...,{\mu _d} \ge 0$ .Thus, the total coding rate of $\bf{X}$  with non-zero mean is 

\begin{equation}
\begin{aligned}
\label{eq18}
{\rm{R}}({\bf{X}}) &= {\rm{R}}({\bf{\bar X}}) + {\rm{R}}({\boldsymbol {\mu }})\\
&= \frac{m}{2}\log \det ({\bf{I}} + \frac{d}{{m{\varepsilon ^2}}}{\bf{\bar X}}{{{\bf{\bar X}}}^T}) + \frac{d}{2}{\log _2}\left( {1 + \frac{{{{\boldsymbol {\mu }}^T}{\boldsymbol {\mu }}}}{{{\varepsilon ^2}}}} \right).    
\end{aligned}
\end{equation}

\subsubsection{correlation between coding rate and data separability}\label{subsubsec2.2.3}

This section discusses the connection between data encoding rate and data separability. Under the condition that the data follows a Gaussian distribution, we prove Theorem 1. Theorem 1 gives the lower bound of the data coding rate and the necessary and sufficient conditions for it to reach the lower bound. This condition illustrates that if and only if every class $\bf{X}$  has the same distribution, the total coding rate of $\bf{X}$  is identical to the sum of $\bf{X}^{j}$ ’s coding rate.

\begin{theorem}[]\label{thm1}
For any $\left\{ {{{\bf{X}}^j} \in {\mathbb{R}^{d \times {m_j}}}} \right\}_{j = 1}^k$  and any $\varepsilon  > 0$ , let  ${\bf{X}} = [{{\bf{x}}_1},...,{{\bf{x}}_m}] = \left[ {{{\bf{X}}^1}, \cdots {{\bf{X}}^k}} \right] \in {\mathbb{R}^{d \times m}}$ with $m = \sum\limits_{j = 1}^k {{m_j}}$ and ${\boldsymbol {\mu }} = \frac{1}{m}\sum\limits_{i = 1}^m {{{\bf{x}}_i} \in {\mathbb{R}^d}}$,then we define the zero mean part ${\bf{\bar X}} = {\bf{X}} - {\boldsymbol {\mu }} \cdot {{\bf{1}}_{1 \times m}}$  .
Let  ${{\bf{X}}^j} = [{\bf{x}}_1^j,...,{\bf{x}}_{{m_j}}^j] \in {\mathbb{R}^{d \times {m_j}}}$ with ${{\boldsymbol {\mu }}^j} = \frac{1}{{{m_j}}}\sum\limits_{i = 1}^{{m_j}} {{\bf{x}}_i^j}  \in {\mathbb{R}^d}$ and ${{{\bf{\bar X}}}^j} = {{\bf{X}}^j} - {{\boldsymbol {\mu }}^j} \cdot {{\bf{1}}_{1 \times {m_j}}}$ . We have

\begin{equation}
\begin{aligned}
\label{eq19}
&\frac{m}{2}\log \det ({\bf{I}} + \frac{d}{{m{\varepsilon ^2}}}{\bf{\bar X}}{{{\bf{\bar X}}}^T}) + \frac{d}{2}{\log _2}\left( {1 + \frac{{{{\boldsymbol{\mu }}^T}{\boldsymbol{\mu }}}}{{{\varepsilon ^2}}}} \right) \\
&\ge \sum\limits_{j = 1}^k {\frac{{{m_j}}}{2}\log \det ({\bf{I}} + \frac{d}{{{m_j}{\varepsilon ^2}}}{{{\bf{\bar X}}}^j}{{({{{\bf{\bar X}}}^j})}^T}) + \frac{d}{{2k}}{{\log }_2}\left( {1 + \frac{{{{({{\boldsymbol{\mu }}^j})}^T}{{\boldsymbol{\mu }}^j}}}{{{\varepsilon ^2}}}} \right)} .    
\end{aligned}
\end{equation}
where the equality holds if and only if
\begin{align}
\frac{{{{{\bf{\bar X}}}^1}{{({{{\bf{\bar X}}}^1})}^T}}}{{{m_1}}} = \frac{{{{{\bf{\bar X}}}^2}{{({{{\bf{\bar X}}}^2})}^T}}}{{{m_2}}} &=  \cdots  = \frac{{{{{\bf{\bar X}}}^k}{{({{{\bf{\bar X}}}^k})}^T}}}{{{m_k}}} = \frac{{{\bf{\bar X}}{{({\bf{\bar X}})}^T}}}{m} \nonumber \\
{{\boldsymbol{\mu }}^1} = {{\boldsymbol{\mu }}^2} &=  \cdots  = {{\boldsymbol{\mu }}^k} = {\boldsymbol{\mu }}.\label{eq20}
\end{align}
\end{theorem}

Since the covariance matrix of the vector ${\bf{x}}_i^j$  is

\begin{equation}
{\sum\nolimits^j} = {\rm E}\left[ {\frac{1}{{{m_j}}}\sum\limits_{i = 1}^{{m_j}} {({\bf{x}}_i^j - {{\bf{\mu }}^j}){{({\bf{x}}_i^j - {{\bf{\mu }}^j})}^T}} } \right] = \frac{{{{{\bf{\bar X}}}^j}{{({{{\bf{\bar X}}}^j})}^T}}}{{{m_j}}}, \label{eq21}   
\end{equation}this means that the lower bound is tight when all the components $\left\{ {{{\bf{X}}^j}} \right\}_{j = 1}^k$  have the same covariance and mean. In particular, for data with a Gaussian distribution, it means that every  ${{\bf{X}}^j}$  has the same distribution, which corresponds to the most inseparable case of data.

The Proof of Theorem 1 is based on the concave property of the $\log \det ( \cdot )$  and $\log ( \cdot )$  functions, and they satisfy Jensen's inequality.

\begin{proof}
	Since  $\log \det ( \cdot )$ and  $\log ( \cdot )$ is strictly concave, The Jensen's inequality is satisfied. We have

\begin{equation}
f(\sum\limits_{j = 1}^k {{\beta _j}} {{\bf{S}}^j}) \ge \sum\limits_{j = 1}^k {{\beta _j}} f({{\bf{S}}^j}). \label{eq22}   
\end{equation}

for all $\left\{ {{\beta _j} > 0} \right\}_{j = 1}^k$ , $\sum\limits_{j = 1}^k {{\beta _j}}  = 1$  and $\left\{ {{{\bf{S}}^j} \in \mathbb{S}_{ +  + }^n} \right\}_{j = 1}^k$ , where equality holds if and only if ${{\bf{S}}^1} = {{\bf{S}}^2} =  \cdots  = {{\bf{S}}^k}$ . 

For function $\log \det ( \cdot )$ , take ${\beta ^j} = \frac{{{m_j}}}{m}$ and ${{\bf{S}}^j} = {\bf{I}} + \frac{d}{{{m_j}{\varepsilon ^2}}}{{{\bf{\bar X}}}^j}{({{{\bf{\bar X}}}^j})^T}$ , we get

\begin{equation}
\log \det ({\bf{I}} + \frac{d}{{m{\varepsilon ^2}}}{\bf{\bar X}}{{{\bf{\bar X}}}^T}) \ge \sum\limits_{j = 1}^k {\frac{{{m_j}}}{m}\log \det ({\bf{I}} + \frac{d}{{{m_j}{\varepsilon ^2}}}{{{\bf{\bar X}}}^j}{{({{{\bf{\bar X}}}^j})}^T})} . \label{eq23}   
\end{equation}
with equality holds if and only if $\frac{{{{{\bf{\bar X}}}^1}{{({{{\bf{\bar X}}}^1})}^T}}}{{{m_1}}} = \frac{{{{{\bf{\bar X}}}^2}{{({{{\bf{\bar X}}}^2})}^T}}}{{{m_2}}} =  \cdots  = \frac{{{{{\bf{\bar X}}}^k}{{({{{\bf{\bar X}}}^k})}^T}}}{{{m_k}}} = \frac{{{\bf{\bar X}}{{({\bf{\bar X}})}^T}}}{m}$.

For function $\log ( \cdot )$ , take ${\beta ^j} = \frac{1}{k}$  and ${{\bf{S}}^j} = 1 + \frac{d}{{{\varepsilon ^2}}}{({{\boldsymbol{\mu }}^j})^T}{{\boldsymbol{\mu }}^j}$ ,we get

\begin{equation}
\log (1 + \frac{{{{\boldsymbol{\mu }}^T}{\boldsymbol{\mu }}}}{{{\varepsilon ^2}}}) \ge \sum\limits_{j = 1}^k {\frac{1}{k}\log } (1 + \frac{{{{({{\boldsymbol{\mu }}^j})}^T}{{\boldsymbol{\mu }}^j}}}{{{\varepsilon ^2}}}). \label{eq24}
\end{equation}
with equality holds if and only if ${{\boldsymbol{\mu }}^1} = {{\boldsymbol{\mu }}^2} =  \cdots  = {{\boldsymbol{\mu }}^k} = {\boldsymbol{\mu }}$  , and the last equality is from $\sum\limits_{j = 1}^k {{m_j}{{\boldsymbol{\mu }}^j} = m} {\boldsymbol{\mu }}$ . 

From formula (\ref{eq23}) and (\ref{eq24}), Theorem \ref{thm1} can be proved.
\end{proof}

\begin{figure}[h]%
\centering
\includegraphics[width=0.9\textwidth]{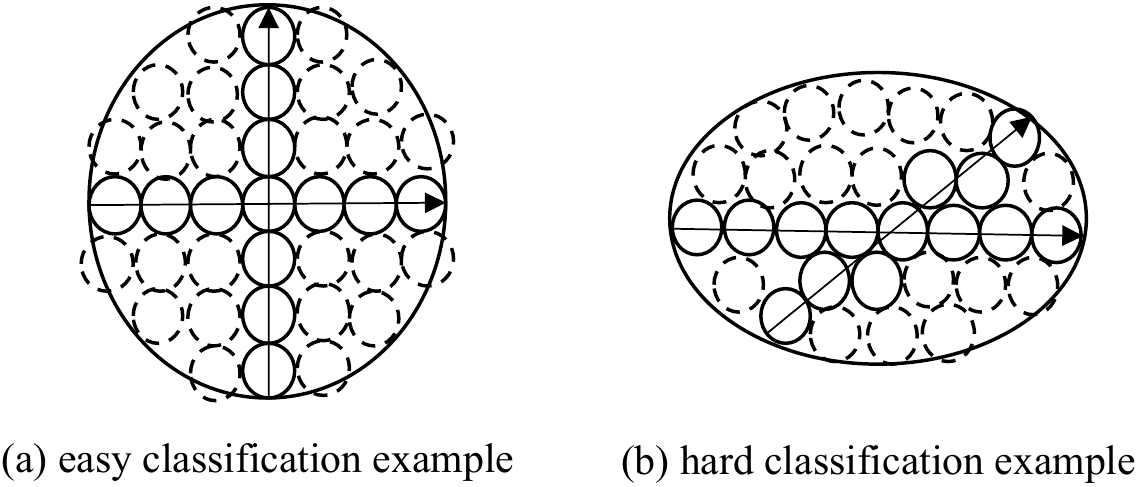}
\caption{Different separability of two datasets. Compared with (b), (a) has a larger total space volume and a smaller intra-class space volume, which seems to have better separability.}\label{fig2}
\end{figure}

 We can now conclude that the sum of the various classes of data coding rate is a lower bound on the overall data coding rate. When the overall data coding rate reaches the lower bound, its necessary and sufficient condition indicate: for the Gaussian distributed data, each category of data has the same distribution, which also means that the feature vectors of each class have a high degree of coincidence, corresponding to the most inseparable situation. In a two-class problem, Figure \ref{fig2} shows the different separability of two datasets and vividly reflects the above point of view. We can suppose that as $\bf{X}$  space expands, each $\bf{X}^{j}$  is compressed, making it easier to classify.
 
According to the preceding analysis, our proposed data separability measure based on rate-distortion is:

\begin{equation}
{{\rm{R}}_{\rm{S}}}({\bf{X}}) = \frac{{{{\rm{R}}_{\rm{C}}}({\bf{X}},\varepsilon \vert{\bf{\Pi }})}}{{{\rm{R}}({\bf{X}},\varepsilon )}}.
\label{eq25}
\end{equation}

Unlike Yu et al. \cite{bib24}, who utilized $\Delta {\rm{R}}({\bf{X}}) = {\rm{R}}({\bf{X}},\varepsilon ) - {{\rm{R}}_{\rm{C}}}({\bf{X}},\varepsilon \vert{\bf{\Pi }})$  as the optimization problem's objective function subjecting to $\left\| {{{\bf{X}}^j}} \right\|_F^2 = {\rm{tr}}({{\bf{\Pi }}^j})$ , here we discard the constraint and adopt a ratio form between ${\rm{R}}({\bf{X}},\varepsilon \vert{\bf{\Pi }})$  and ${\rm{R}}({\bf{X}},\varepsilon )$ , resulting in a data separability measure ${{\rm{R}}_{\rm{S}}}({\bf{X}})$  in the range of [0,1] with low values indicating high separability.

\section{Experiments}\label{sec3}

We first verify the proposed measure RS's effectiveness using a two-class synthetic dataset\footnote{These datasets are created by the Samples Generator in
sklearn.datasets 
\url{https://scikit-learn.org/stable/modules/classes.html\#samples-generator}} with adjustable separability, and contrast its separability evaluation results with DSI, N2, LSC, and Density.

\subsection{datasets with different feature shapes}\label{subsec3.1}

\begin{figure}[h]%
\centering
\includegraphics[width=0.9\textwidth]{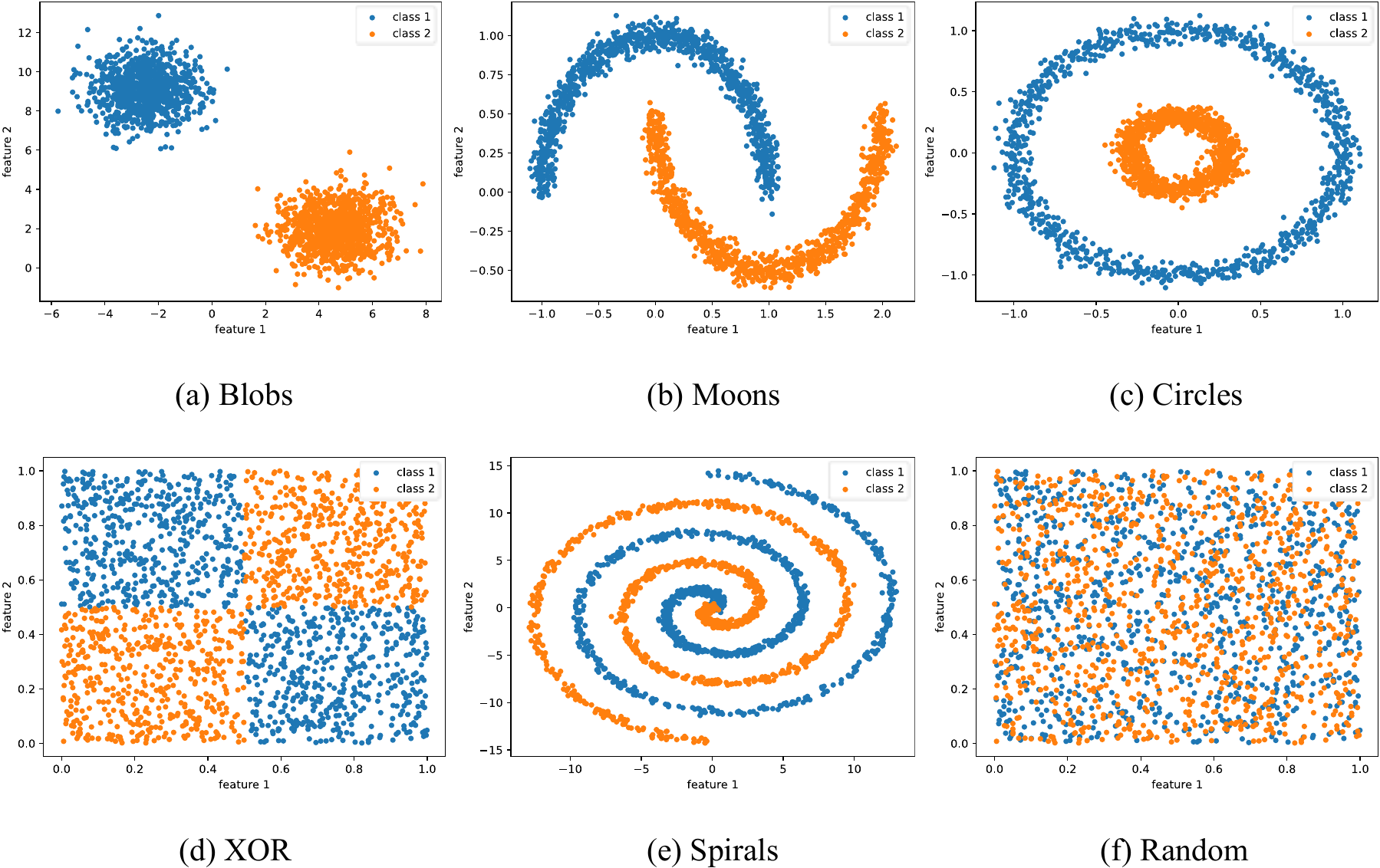}
\caption{Typical two-class datasets: separability is decreasing from Figure (a) to Figure (f).}\label{fig3}
\end{figure}

The most basic type of data separability analysis is the linear separability of data \cite{bib25}. Linear separability for a two-class dataset is defined as: There exists a hyperplane, which can correctly divide the positive and negative instance points to both sides of the hyperplane. An effective measure must first be capable of measuring data linear separability. Therefore, we created two-class datasets with six distinct feature shapes, showing clearly discernible linear separability. Each class consists 1000 samples of 2 dimensions. Figure \ref{fig3} displays datasets feature plots and Table 1 lists the measures’ performance.

According to Table \ref{tab1}, we can conclude that RS can correctly represent the relative gap of linear separability among datasets while maintaining excellent comparability accuracy to four decimal places. In general, the other measures can also roughly reflect relative separability. But in terms of details, DSI deviated from the separability analysis of the Circle data. When assessing the Random data, N2 has increased significantly. LSC confuses the separability of XOR and Spirals, And the Density's evaluation results of Circle, XOR, and Spirals are unclear.

\begin{table}[h]
\begin{center}
\begin{minipage}{\textwidth}
\caption{Separability measures results for the two-class datasets (Figure \ref{fig3})}\label{tab1}%
\begin{tabular*}{\textwidth}{@{\extracolsep{\fill}}cccccccc@{\extracolsep{\fill}}}
\toprule
&Measures & Blobs  & Mooms & Circles & XOR & Spirals & Random\\
\midrule
&DSI    & 0.0008   & 0.6413  & 0.4483 & 0.7775 & 0.9468 & 0.9992 \\
&N2    & 0.0007   & 0.0014  & 0.0011 & 0.0127 & 0.0789 & 0.8140 \\
&LSC    & 0.5137   & 0.9083  & 0.8400 & 0.9997 & 0.9997 & 0.9999 \\
&Density & 0.0758   & 0.6707   & 0.5580  & 0.7714 & 0.7406 & 0.8565 \\
&RS(Proposed)  & 0.1402 & 0.6560 & 0.8571 & 0.8700 & 0.9847 & 0.9982  \\
\botrule
\end{tabular*}
\footnotetext{A low value indicates high separability.}
\end{minipage}
\end{center}
\end{table}
\subsection{datasets with different overlap regions }\label{subsec3.2}

\begin{figure}[H]%
\centering
\includegraphics[width=0.9\textwidth]{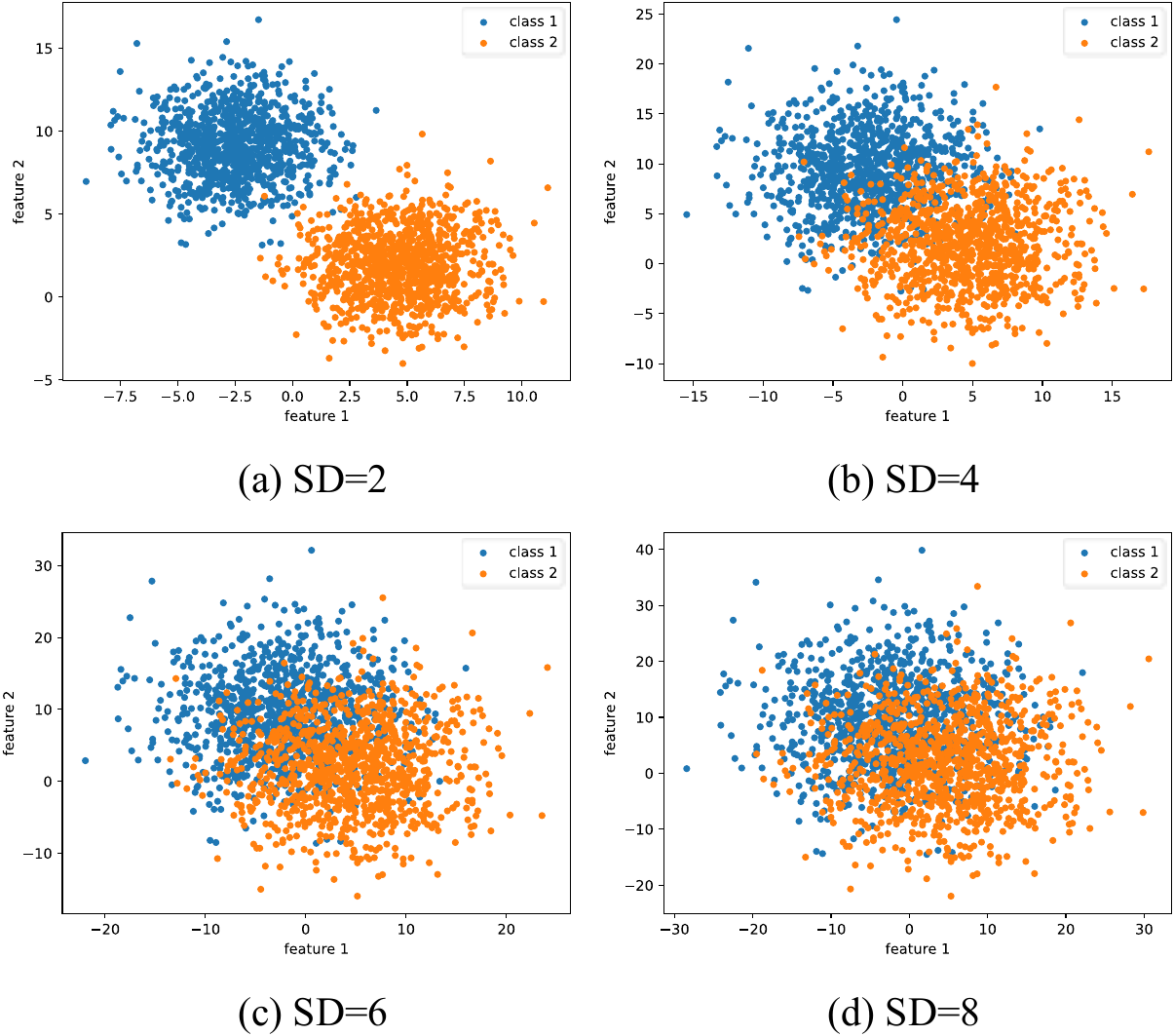}
\caption{The Blobs data with different cluster standard deviations (SD). A high SD value denotes a significant overlap area.}\label{fig4}
\end{figure}

Now consider another case in which the feature distribution exhibits a high degree of coincidence, making it challenging to categorize the data. Can data separability indicators measure this situation? To answer the question, we experiment on the Blobs data. Since the features of the Blobs data follow a Gaussian distribution, the region of feature overlap can be adjusted by changing the feature standard deviation (SD). We set the SD parameter from 1 to 9; four instances are depicted in Figure \ref{fig4}.

\begin{figure}[h]%
\centering
\includegraphics[width=0.9\textwidth]{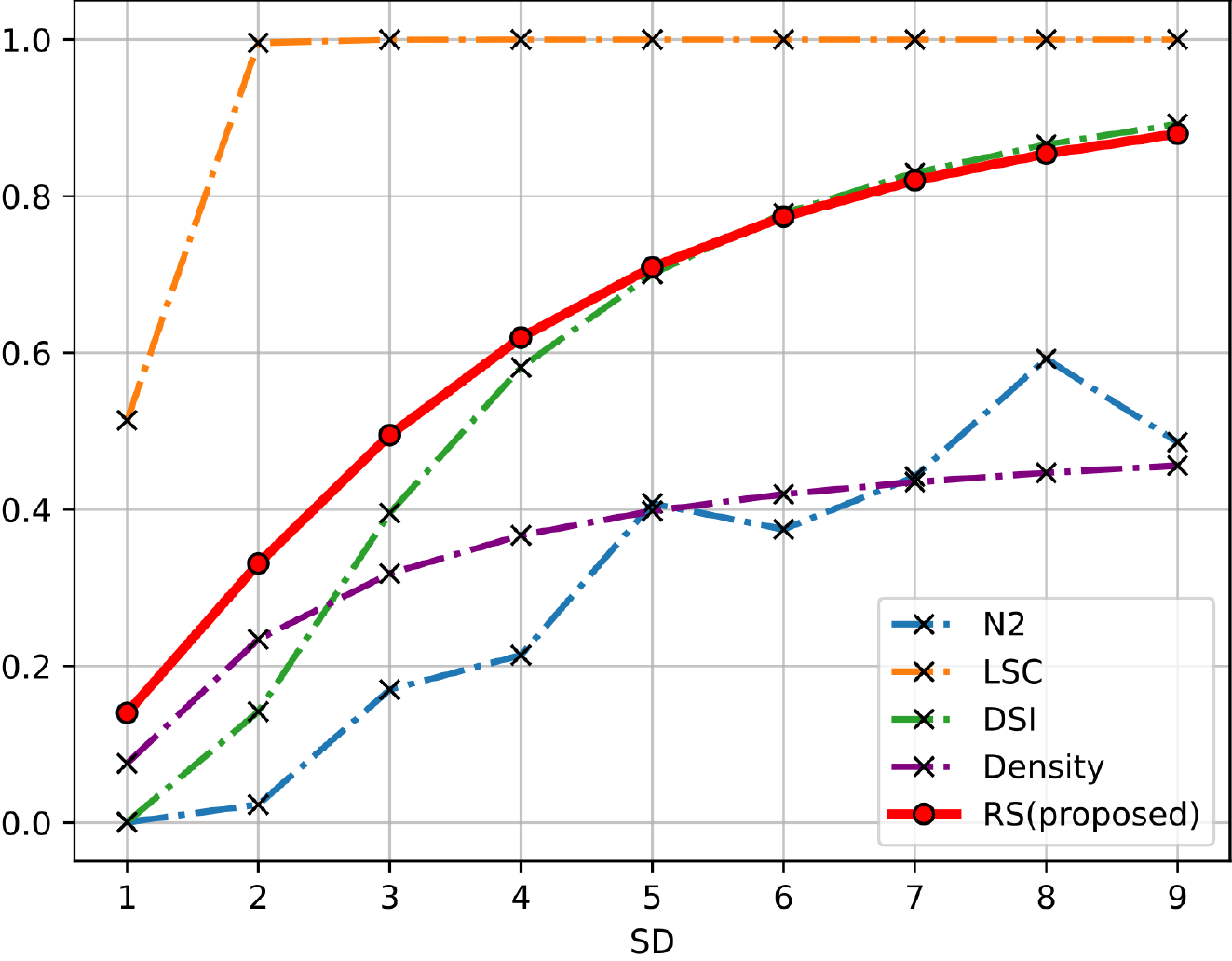}
\caption{Comparison of data separability evaluation results with varying degrees of feature overlap. For measures, a high value on the y-axis indicates low separability.}\label{fig5}
\end{figure}

Then, for the nine Blobs datasets, we compute DSI, N2, LSC, Density, and the proposed measure (RS) and present them together with SVM classification accuracy as a baseline for separability in Figure \ref{fig5}.

In this condition, both N2 and LSC fail to assess the data separability. Among them, LSC can only distinguish the case of features with or without overlap and is not sensitive to the change of feature overlap area. At the same time, N2 fluctuates with the deterioration of data separability, suggesting a lower evaluation precision. Besides, RS, DSI, and Density can correctly reflect the trend of data separability, i.e., a high SD value corresponds to a high measure value. Furthermore, both DSI and RS have an extensive dynamic change range.

\subsection{datasets with different preprocessing methods}\label{subsec3.3}

Feature preprocessing is one of the most prevalent approaches to ensure the classification performance of the classifier. Some types of preprocessing are employed before feeding data to the classifier or before the network extracts latent features from the data \cite{bib26}. Therefore, the classifier's data is often preprocessed data, not the original data. The purpose of this experiment is to examine the impact of preprocessing operations on the performance of the separability measure. Take the Blobs data (SD=1) as an example; Table \ref{tab2} summarizes the common feature preprocessing methods.

\begin{table}[h]
\begin{center}
\begin{minipage}{\textwidth}
\caption{Feature preprocessing methods}\label{tab2}%
\begin{tabular*}{\textwidth}{@{\extracolsep{\fill}}cccc@{\extracolsep{\fill}}}
\toprule
&Preprocessing methods & Operations  &  Examples\\
\midrule
&Min-max normalization    
& ${\bf{x}} \leftarrow \frac{{{\bf{x}} - {\rm{min}}({\bf{x}})}}{{\max ({\bf{x}}) - \min ({\bf{x}})}}$   
& \begin{minipage}[b]{0.3\columnwidth}
		\raisebox{-.5\height}{\includegraphics[width=\linewidth]{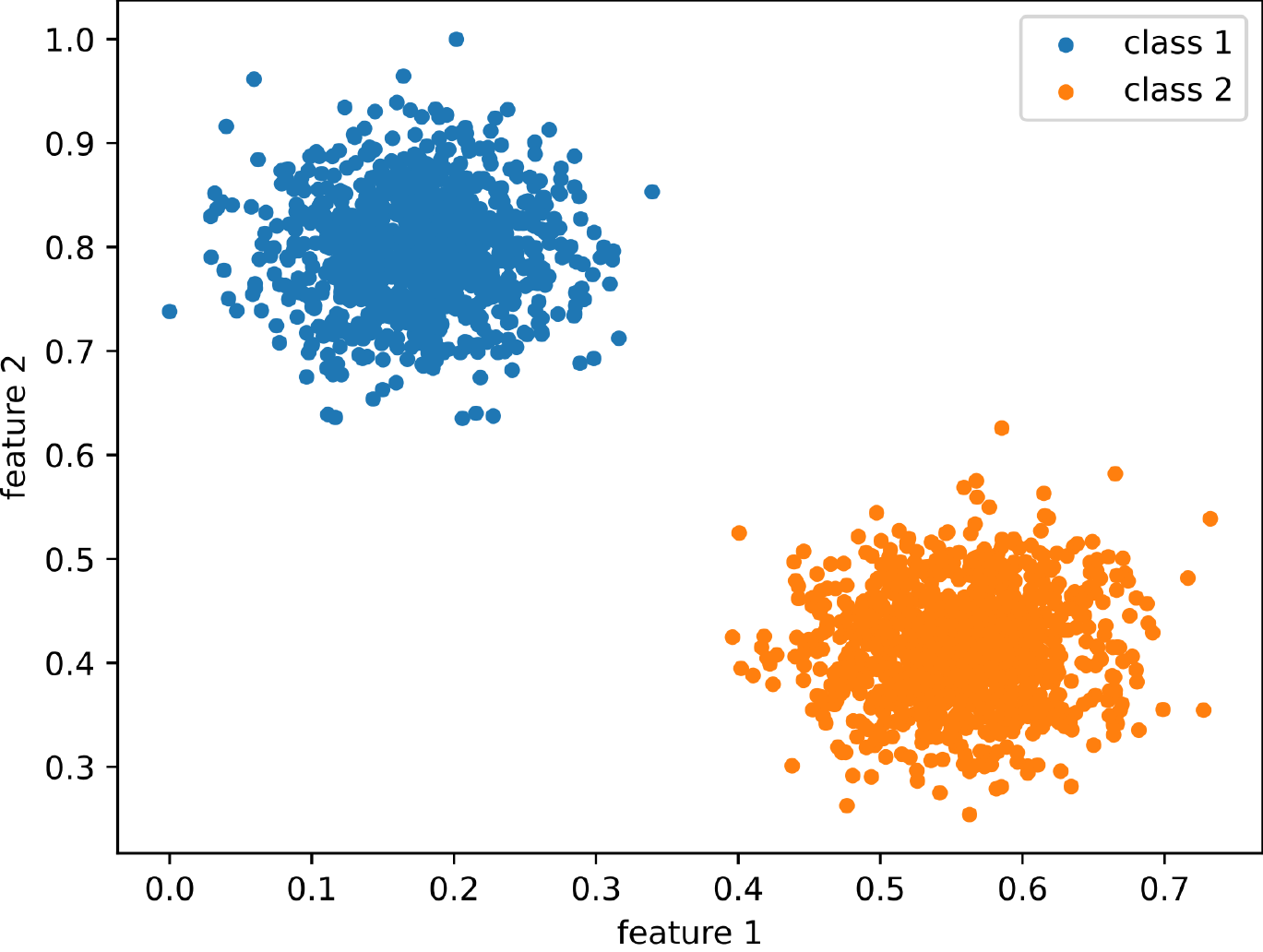}}
	\end{minipage} \\
&Mean normalization    
& ${\bf{x}} \leftarrow \frac{{{\bf{x}} - {\rm{mean}}({\bf{x}})}}{{\max ({\bf{x}}) - \min ({\bf{x}})}}$   
& \begin{minipage}[b]{0.3\columnwidth}
		\raisebox{-.5\height}{\includegraphics[width=\linewidth]{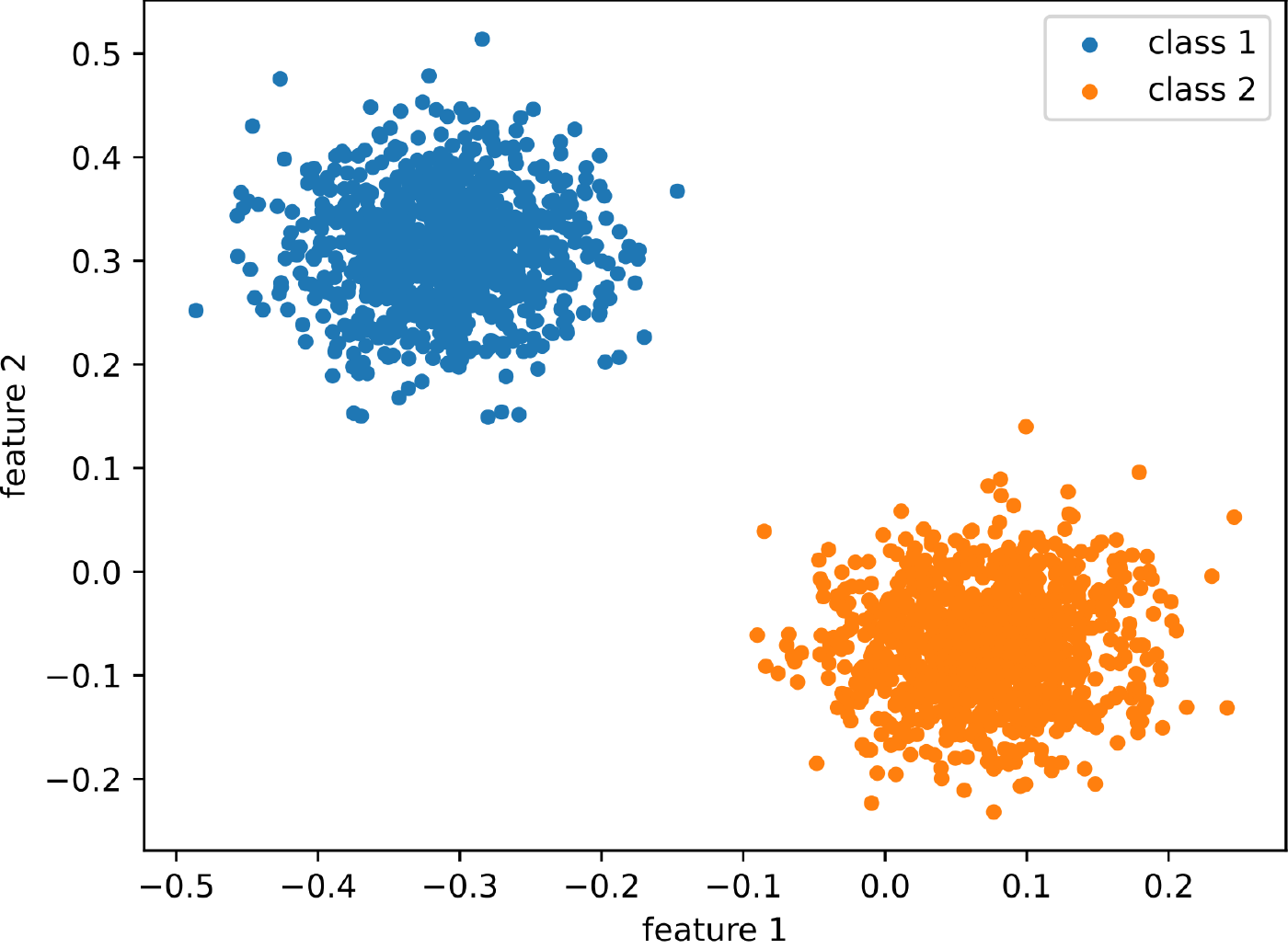}}
	\end{minipage} \\
&L2 normalization    
& ${\bf{x}} \leftarrow \frac{{\bf{x}}}{{\vert \vert{\bf{x}}\vert{\vert_2}}}$   
& \begin{minipage}[b]{0.3\columnwidth}
		\raisebox{-.5\height}{\includegraphics[width=\linewidth]{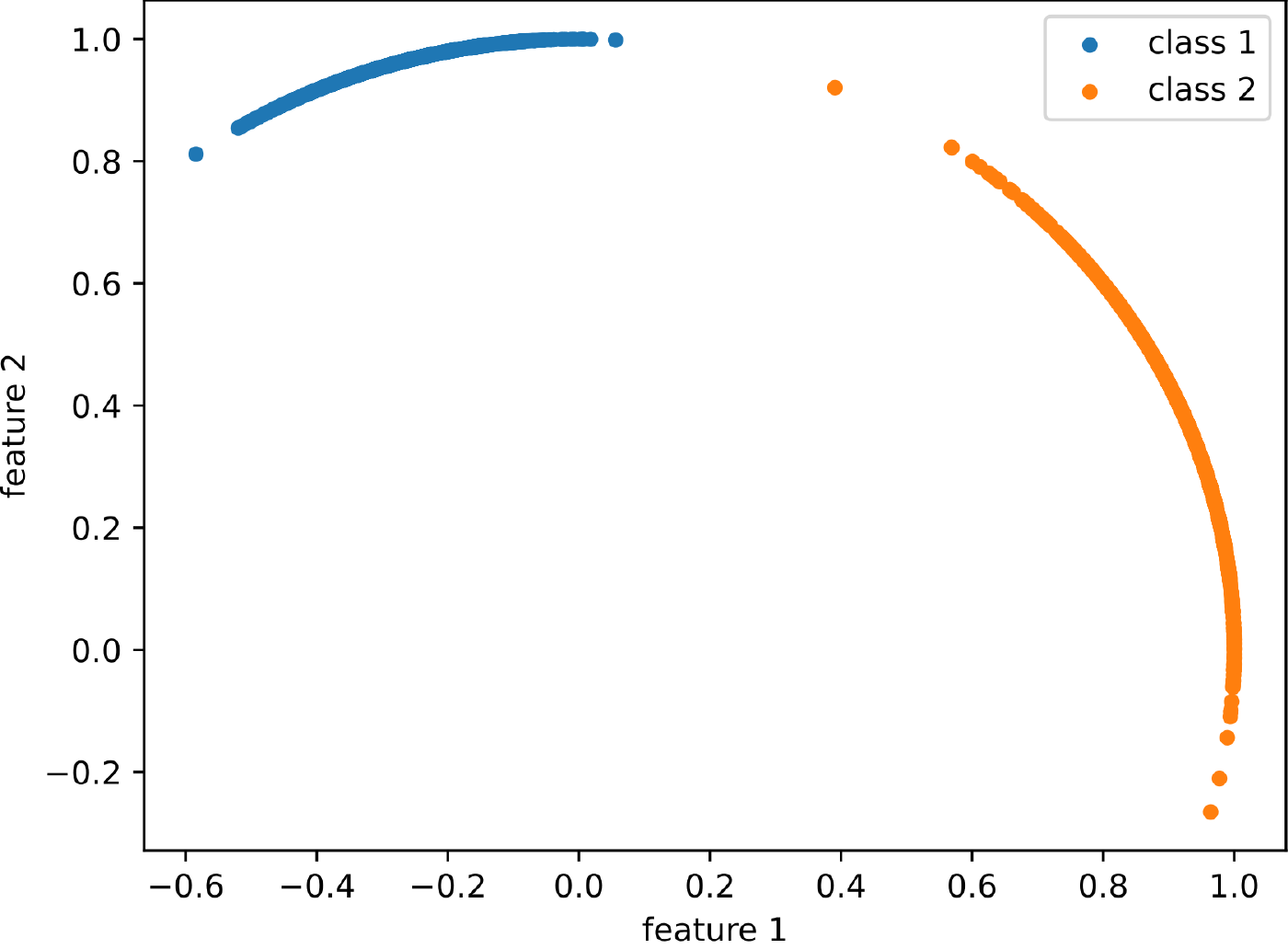}}
	\end{minipage} \\
&Centralization   
& ${\bf{x}} \leftarrow {\bf{x}} - {\rm{mean}}({\bf{x}})$   
& \begin{minipage}[b]{0.3\columnwidth}
		\raisebox{-.5\height}{\includegraphics[width=\linewidth]{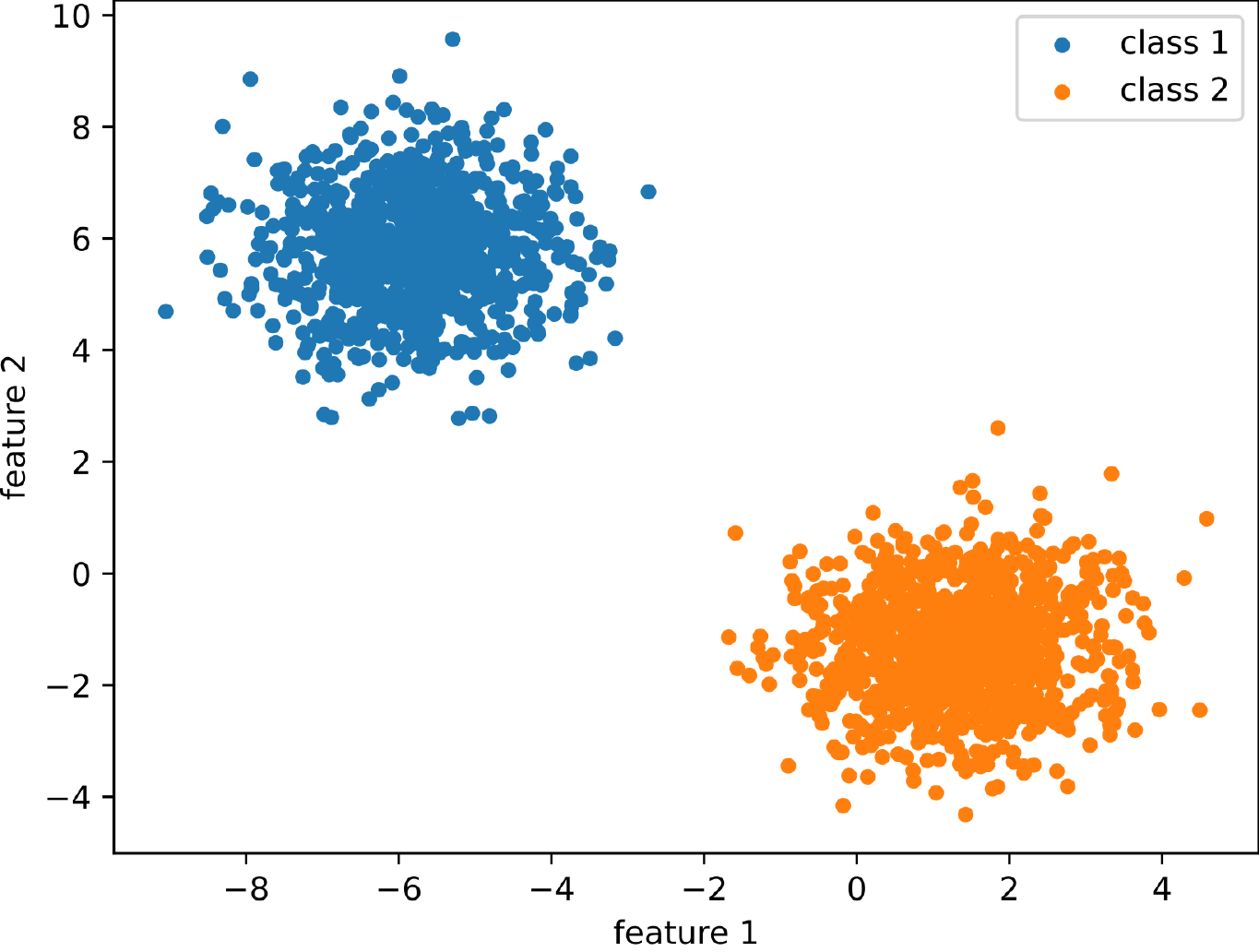}}
	\end{minipage} \\
&Standardization   
& ${\bf{x}} \leftarrow \frac{{{\bf{x}} - {\rm{mean}}({\bf{x}})}}{{{\rm{std}}({\bf{x}})}}$   
& \begin{minipage}[b]{0.3\columnwidth}
		\raisebox{-.5\height}{\includegraphics[width=\linewidth]{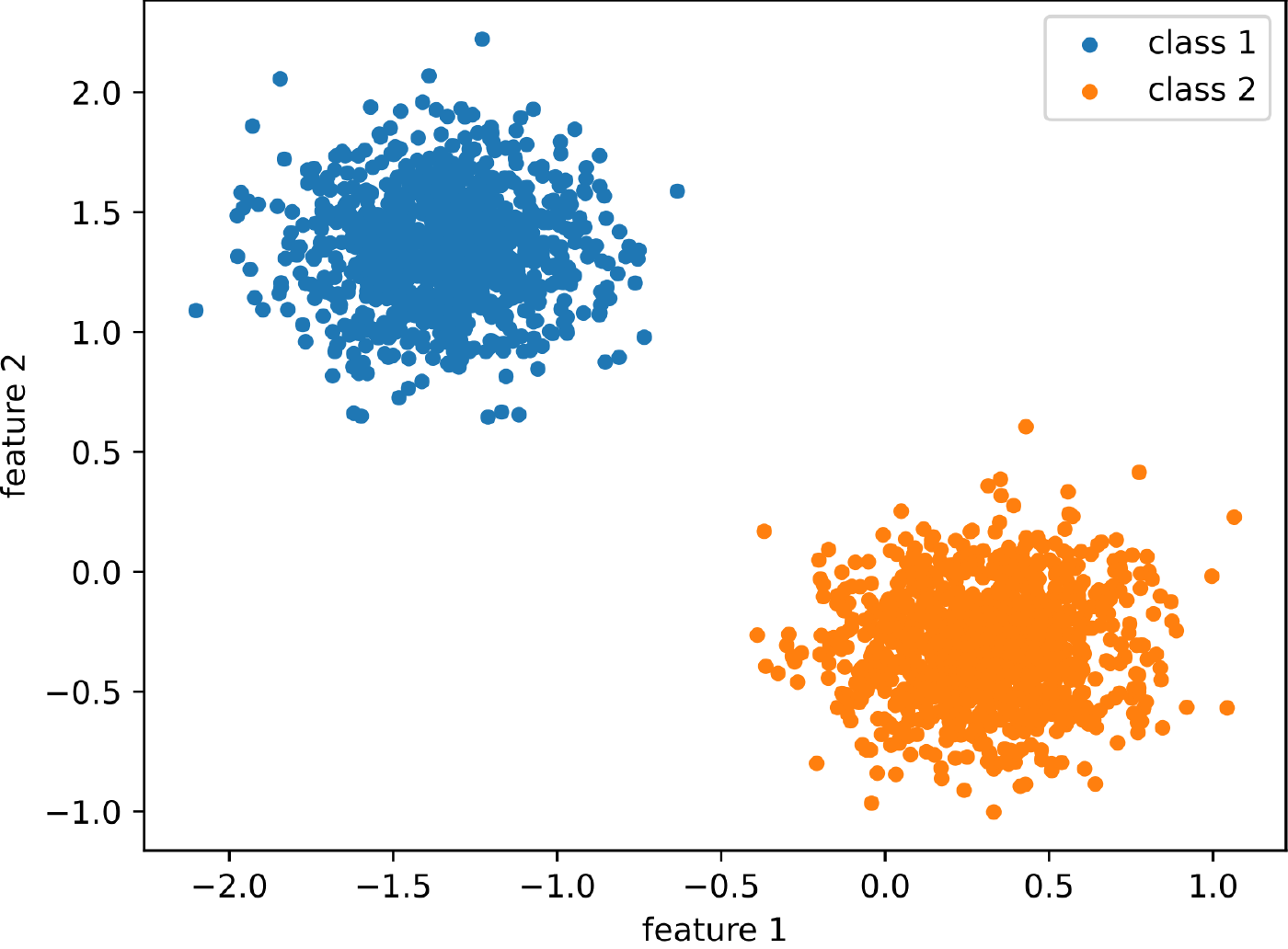}}
	\end{minipage} \\
\botrule
\end{tabular*}
\end{minipage}
\end{center}
\end{table}

\begin{figure}[h]%
	\centering
	\includegraphics[width=0.9\textwidth]{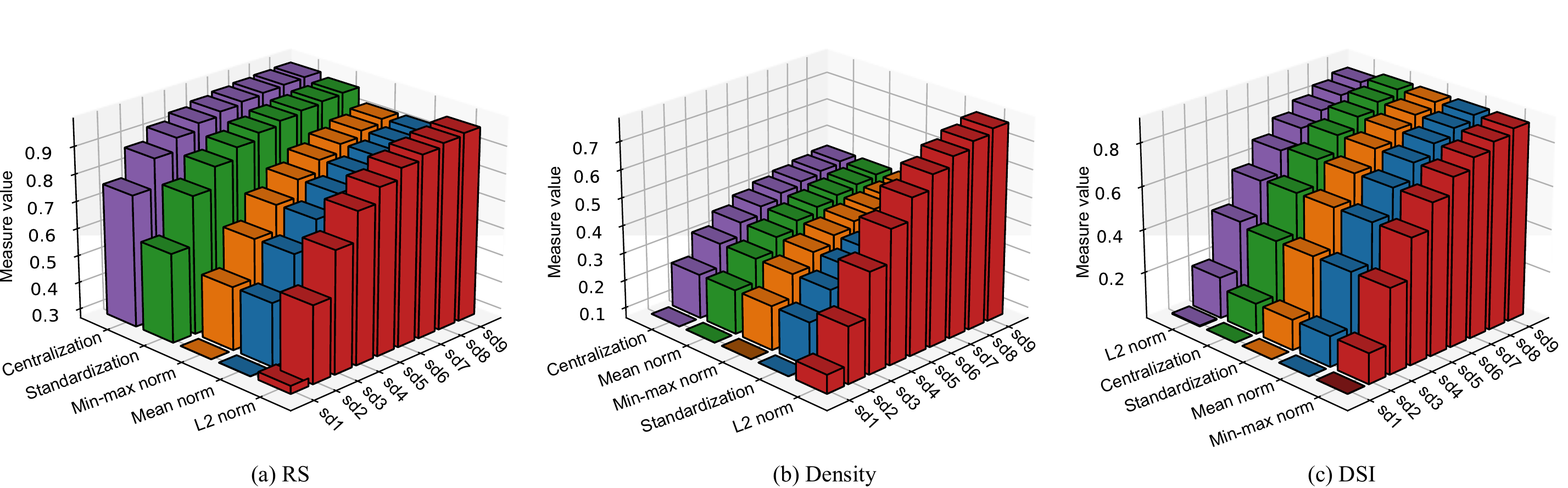}
	\caption{Sensitivity test of the measures to preprocessing methods. In each figure, the groups of preprocessing methods are sorted according to the variance of the measure values corresponding to each group from front to back.}\label{fig6}
\end{figure}

From Table \ref{tab2}, we can see that the Min-max, Mean normalization, Centralization, and Standardization change the range of feature values without changing the cluster shape. L2 normalization differs in that it alters the feature shape by normalizing the vector length to a unit circle. Except for L2 normalization, the data after each preprocessing shows no difference in separability from the feature maps. Hence, we conjecture that data preprocessing does not affect the separability of the data. So are the separability measures sensitive to the preprocessing methods? We tested RS, DSI, and density on the Blob (SD = 1, 2,...,9), and the results are shown in Figure \ref{fig6}.

Figure \ref{fig6} presents commonalities that RS, Density, and DSI can all indicate the relative gap in the separability of each dataset when exposed to various preprocessing methods. Yet the influence of preprocessing techniques on measures performance is different. With the same data separability (fixed SD), the value of RS is more sensitive to the preprocessing methods. It has the most significant variance in the L2 normalization group as the SD value increases. Density also exhibits a greater range of dynamic variation on the L2-normalization. DSI shows strong consistency in all preprocessing methods. Consequently, we use the preprocessing process of L2 normalization without modifying data separability to improve the proposed measure of RS's evaluation performance. 

\subsection{datasets with different complexity decision boundaries}\label{subsec3.4}

\begin{figure}[H]%
	\centering
	\includegraphics[width=0.9\textwidth]{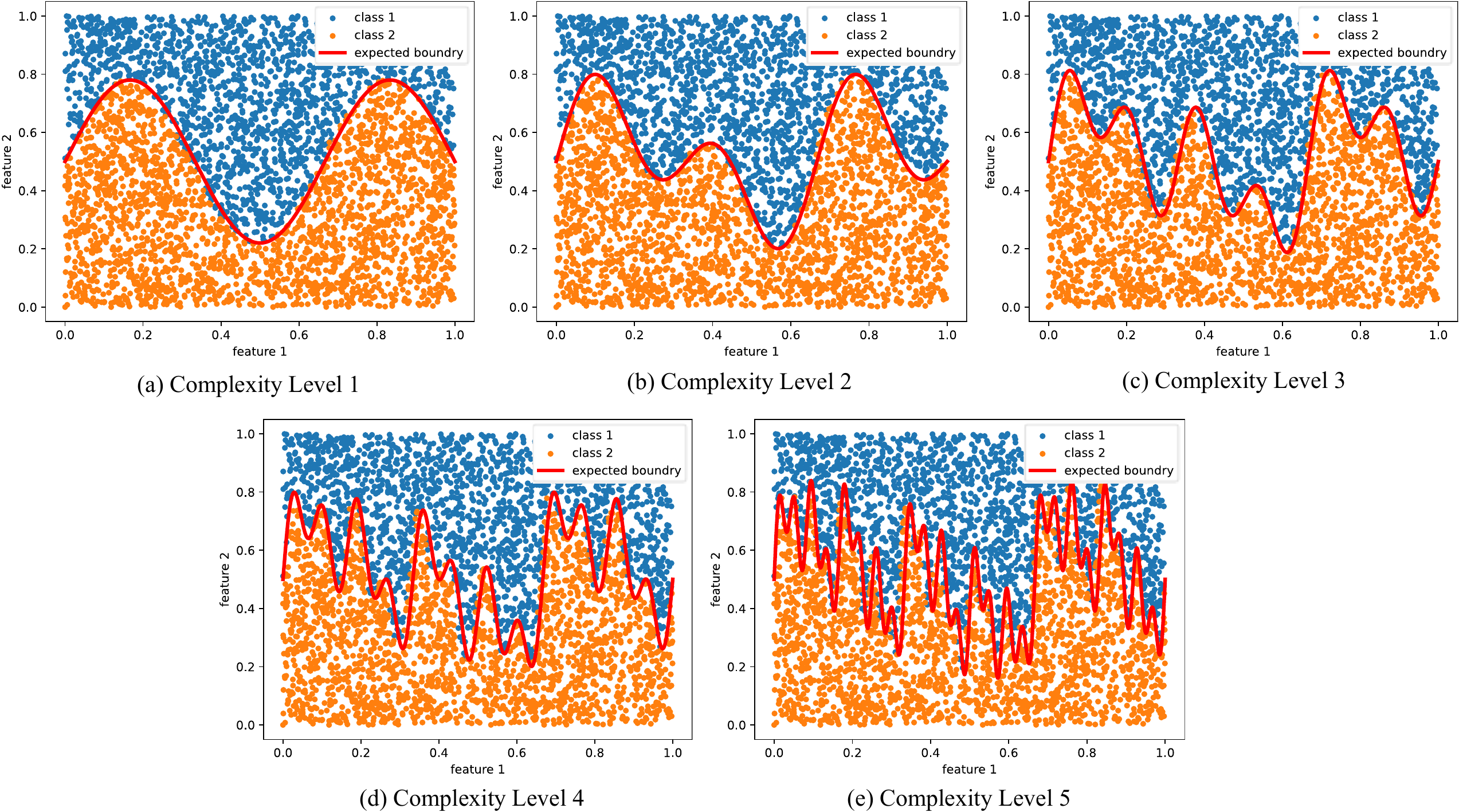}
	\caption{The datasets with different complexity decision boundaries.}\label{fig7}
\end{figure}
Including the constraint situation Guan et al. mentioned \cite{bib20}, we last examine RS’s performance when measuring the complexity of data points on or near the decision boundary. In this experiment, we construct a two-class dataset with 2000 samples of 2 features for each class. The feature maps and the expected decision boundaries are shown in Figure \ref{fig7}.

\begin{figure}[h]%
\centering
\includegraphics[width=0.9\textwidth]{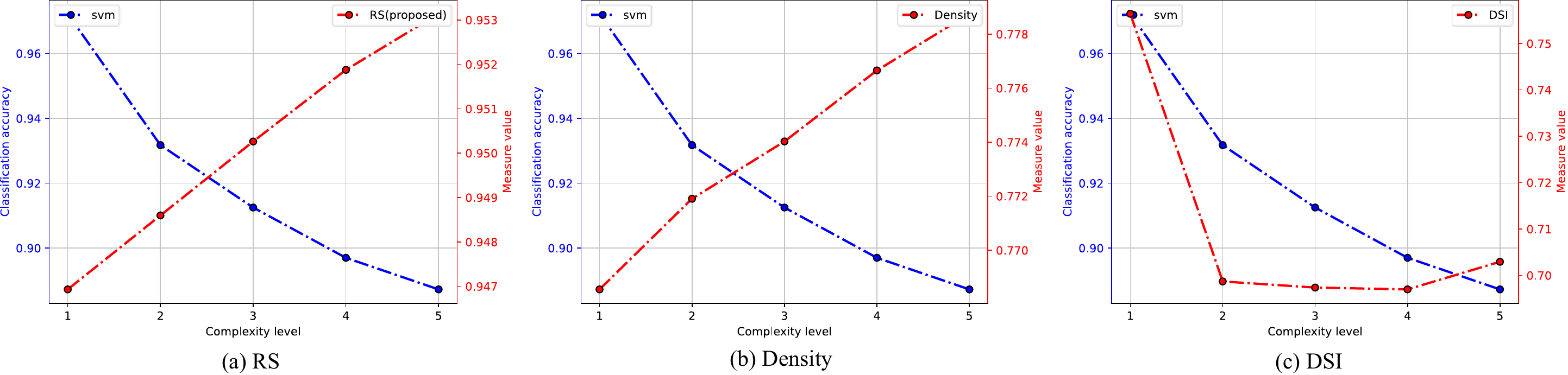}
\caption{The complexity of boundary evaluation results. High measure value and low classification accuracy correspond to high complexity levels.}\label{fig8}
\end{figure}

We use SVM classification accuracy to reflect classification difficulty. Figure \ref{fig8} compares the evaluation performance of RS, Density, and DSI on the datasets (Figure \ref{fig7}).
	
Figure \ref{fig8} shows that only DSI fails, but RS and Density achieve the necessary progress, a growing trend as complexity levels. Although the increment of measure value is slight with a whole range of 0.006, it explains that the tiny change in RS value is meaningful.

\section{Discussion}\label{sec4}

After thoroughly verifying the effectiveness of RS, we apply it to real noisy datasets with unknown structures. Considering that the recognition accuracy of the classifier on the dataset can reflect the complexity of real data to a certain extent, we use three machine learning classifiers as additional verification tools: SVM, SGD, and KNN. Classifiers are tested in a noisy environment, i.e., the original dataset is retained as training data, and Gaussian white noise with varying variances is added to it to generate a series of recognition tasks as test data. Eventually, we confirm the correlation between the recognition accuracy of the classifier in the recognition task and the separability evaluation results of RS on noisy data.

Classifiers' accuracy indicates data's separability. And try another perspective, since the classifier is trained and tested based on data, its recognition accuracy essentially relies on how good the data separability is. The part of the discussion is intended to explain this dependence. Similarly, using a real noisy dataset, we investigate the relationship between data separability and the trend of classifier generalization ability in noisy environments. The key is to quantify the classifier's generalization ability. Specifically, under the premise of meeting certain assumptions, we construct a mapping model from the known classification accuracy to the classifier's anti-noise ability, using RS as the parameter. Furthermore, the mapping model can evaluate the anti-noise performance of particular classifiers within certain limits.

In the last part, we discuss the deep learning model recognition performance evaluation method. As is known, increasing the depth of a neural network improves its ability to extract high-dimensional features. But so far, it is difficult to explain the substantive role of each module in the network. Consequently, we consider a modular evaluation method. The separability measure RS, in particular, is used to judge whether each convolution module works by evaluating the variation in the separability of the features extracted by each convolution layer in the deep convolution neural network (CNN).

\subsection{Correlation between classification accuracy and data separability}\label{sec4.1}

\begin{figure}[h]%
	\centering
	\includegraphics[width=0.9\textwidth]{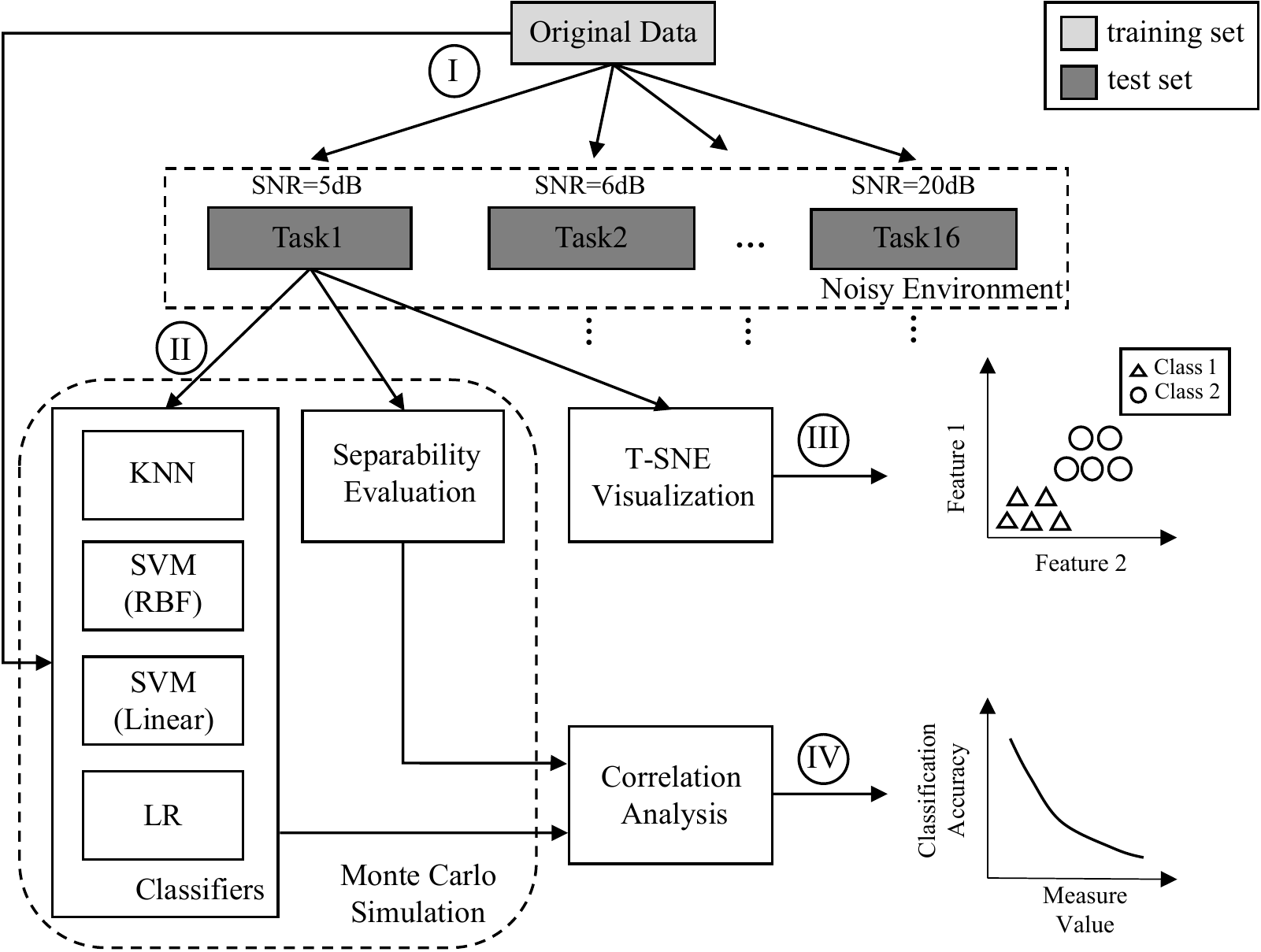}
	\caption{Experiment procedure to verify the correlation between classification accuracy and data separability.}\label{fig9}
\end{figure}

After thoroughly verifying the validity of RS as a measure of data separability, we characterize the data separability using RS values. This section discusses the experimental procedures used to investigate the correlation between classification accuracy and data separability. In this study, a noisy environment with 16 sets of recognition tasks was designed based on real data. And to verify the rationality of constructed scenario, the T-SNE tool is used to visualize the influence of noise on data separability. And then, with the original data as a training set and the noisy data as a test set, the recognition accuracy of the four classifiers in noisy environments is obtained. Next, we compute noisy data's RS value and analyze the classification accuracy correlation. Here we apply the Monte Carlo simulation to average the randomness of the results due to noise. The experiment framework is shown in Figure \ref{fig9}.

Step I is to add Gaussian white noise with a specific variance to the original data to create a test set with a signal-to-noise ratio (SNR) of 5-20 dB, and the formula involving the SNR computation is equation (\ref{eq26})

\begin{equation}
	SNR = 10\lg \frac{{{P_{\bf{X}}}}}{N} = 10\lg \frac{{\sum\limits_{d = 1}^D {{P_d}} }}{N}, \label{eq26}   
\end{equation}

where ${P_{\bf{X}}}$  is the average power of the original data ${\bf{X}}$ , $N$  is the noise's average power, $D$  is the ${\bf{X}}$ 's feature dimension, and ${P_d}$  is the power on the $d$ dimension.

In Step II, we deploy four standard machine learning classifiers. Nonlinear classifiers such as K-Nearest Neighbor (KNN) and Support Vector Machine with Radial Basis Function (SVM with RBF) can generate nonlinear decision boundaries. Linear classifiers include linear SVM and logistic regression (LR). The UCI datasets \cite{bib27} are utilized in the experiment. Table \ref{tab3} and \ref{tab4} present the detailed parameter setting of the classifiers and datasets\footnote{The URL for downloading the dataset:\url{https://archive.ics.uci.edu/}}.

 \begin{figure}[h]%
	\centering
	\includegraphics[width=0.9\textwidth]{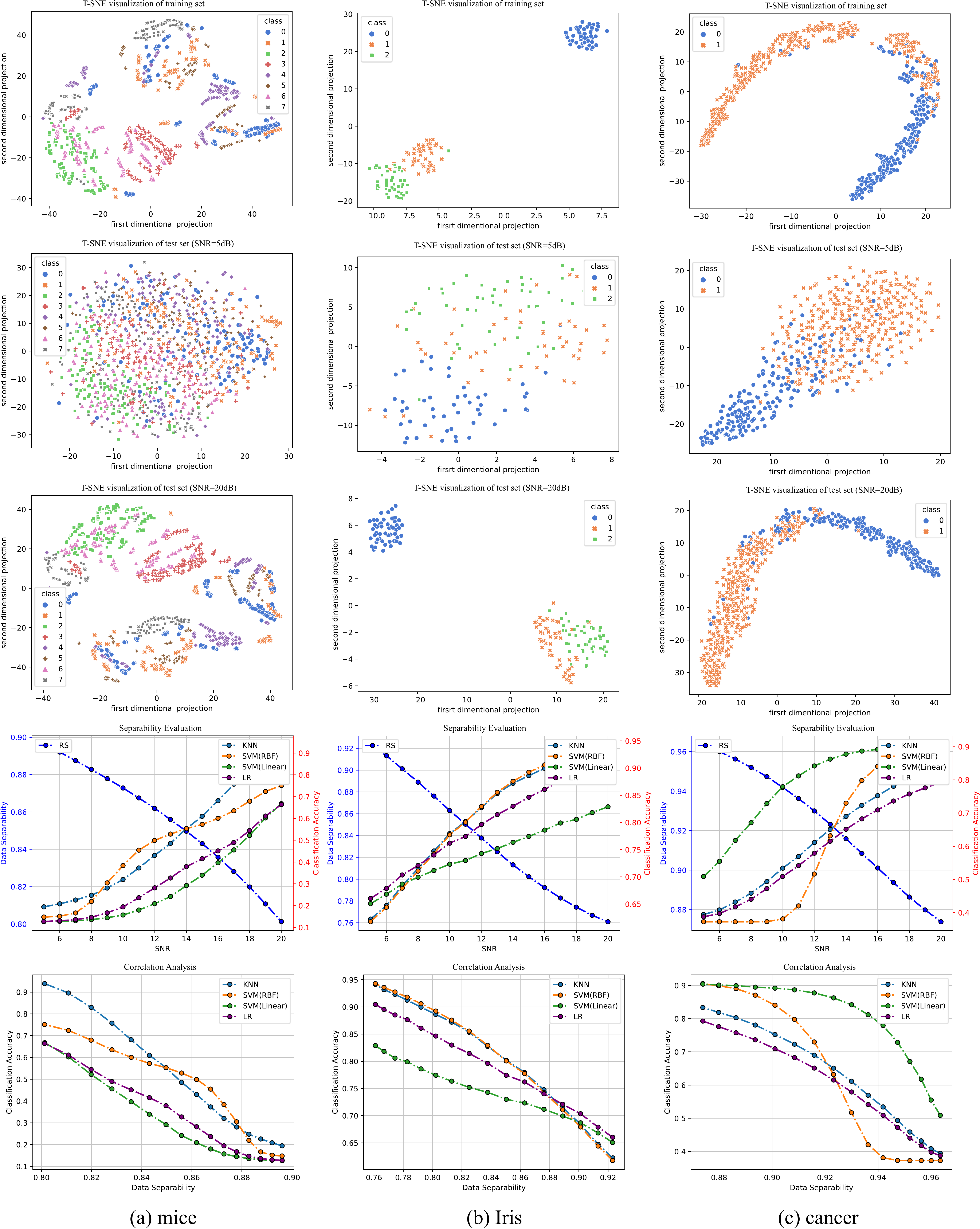}
	\caption{Analysis results to verify the correlation between classification accuracy and data separability.}\label{fig10}
\end{figure}

\begin{table}[h]
	\begin{center}
		\begin{minipage}{\textwidth}
			\caption{Machine learning models}\label{tab3}%
			\begin{tabular*}{\textwidth}{@{\extracolsep{\fill}}ccc@{\extracolsep{\fill}}}
				\toprule
				&Classifiers & Parameter setting  \\
				\midrule
				&KNN     & number of nearest neighbor nodes  $K = 5$    \\
				&SVM (RBF)    & regular coefficient $C = 1$    \\
				&SVM (Linear)    & regular coefficient $C = 1$    \\
				&LR & regular coefficient $\alpha  = 0.0001$    \\
				\botrule
			\end{tabular*}
		\end{minipage}
	\end{center}
\end{table}

\begin{table}[h]
	\begin{center}
		\begin{minipage}{\textwidth}
			\caption{UCI datasets}\label{tab4}%
			\begin{tabular*}{\textwidth}{@{\extracolsep{\fill}}ccccc@{\extracolsep{\fill}}}
				\toprule
				& Datasets &Instances  & Features & Classes \\
				\midrule
				&  Mice	 & 1080 & 77 & 8  \\
				& Iris & 150 & 4 & 3  \\
				& Cancer & 569 & 30 & 2 \\
				\botrule
			\end{tabular*}
		\end{minipage}
	\end{center}
\end{table}

Steps III and IV submit the final analysis results, as seen in Figure \ref{fig10}.

Since real data are usually high-dimensional, we apply the T-SNE tool to observe the variation of data separability with SNR. When the SNR is lower than 5 dB, the test data separability becomes extremely poor relative to the training data. In this condition, all kinds of sample points are scattered, the volume of the subspace in which they are located is expanded, and the proportion of the coding rate of the whole data space is increased. This phenomenon also well corresponds to our definition of RS. When the SNR is high as 20dB, its separability is equivalent to the training data, and the two have almost the same feature distribution. Referring to the separability evaluation and correlation analysis, as SNR grows, the separability of test data gradually improves, and the classification accuracy increases along with it. Finally, we can conclude a strong positive correlation between data separability and recognition accuracy.

\subsection{Classifier’s ability evaluated by classification accuracy under data separability}\label{sec4.2}

We verify the positive correlation between classification accuracy and data separability in the previous discussion. For the same type of classifier with fixed parameters, the better the separability of the test data provided by the recognition task, the higher the recognition accuracy the classifier exhibits. For different classifiers, we can explicitly compare the recognition accuracy of each classifier on an unknown separability recognition task, implying that each classifier's recognition ability is directly given by its recognition accuracy. This conclusion applies only to the circumstance of classifier recognition performance evaluation under a single task. But in a realistic scenario, the classifier is faced with a set of recognition tasks in a complex recognition environment as constructed in Figure \ref{fig9}, and we indeed want to evaluate the classifier's generalization ability in this group of tasks. Specifically, as shown in Figure \ref{fig10}(b), for the dataset Iris, the classification accuracy of SVM (RBF) is consistently higher than other classifiers at SNR=20dB, but the lowest at SNR=5dB. Since the 5dB task is more difficult than the 20dB one, we can't conclude whether the classifier performance is good or not. At this point, how could the classifier's performance be measured?

The simple idea is to assign a certain weight ${\bf{W}} \in {\mathbb{R}^n}$  to the recognition accuracy ${{\bf{P}}_{acc}} \in {\mathbb{R}^n}$  of the classifier on that group of tasks according to the difficulty of the recognition task, and $n$  is the number of tasks. The classification ability $\theta $  on these tasks is defined as

\begin{equation}
	\theta  = {{\bf{W}}^T}{{\bf{P}}_{acc}}.\label{eq27}
\end{equation}

${\bf{W}}$ is determined by the difficulty of the recognition task. The more difficult the task, the higher the weight value. According to the prior experiment, the task difficulty depends to some extent on the data separability. Thus, ${\bf{W}}$  as a mapping matrix is parameterized by the separability ${{\bf{R}}_{\rm{S}}}$ . To quantify this mapping relationship, we seek a functional form $f( \cdot )$  of the mapping matrix ${\bf{W}}$.

\begin{equation}
	\theta  = f({P_{acc}};{R_{\rm{S}}})\label{eq28}
\end{equation}

$f( \cdot )$ needs to be obtained by fitting a given ${P_{acc}}$ and $\theta$ . ${P_{acc}}$  can be derived directly from the classification results, whereas $\theta$  is uncertain. Therefore, it is first necessary to construct the known $\theta$ based on the following assumptions.

1) For different difficulty tasks, homogeneous classifiers with fixed parameters exhibit different recognition accuracies.

2) For the same dataset, homogeneous classifiers with fixed parameters exhibit consistent recognition ability values.

3) For homogeneous classifiers with different parameter settings, their relative ability value can be inferred from the recognition accuracy.

\begin{figure}[h]%
	\centering
	\includegraphics[width=0.9\textwidth]{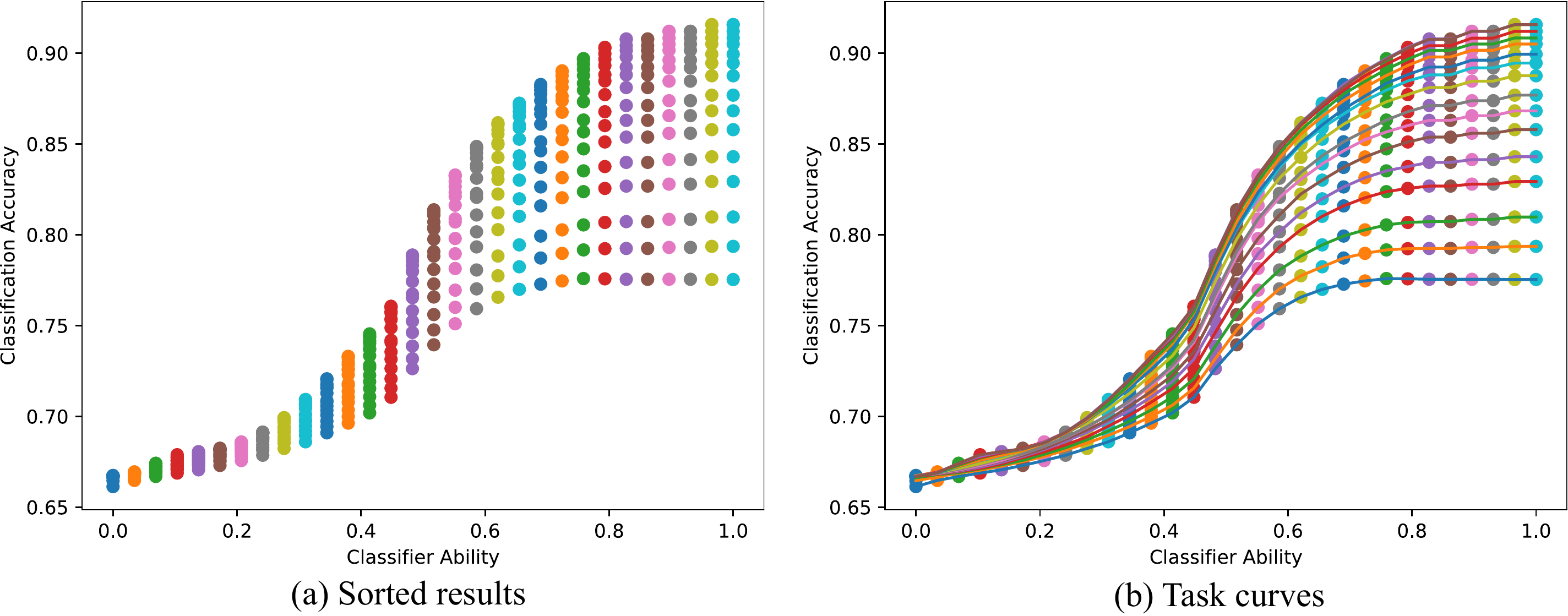}
	\caption{The map of ${{\bf{P}}_{{\rm{acc}}}}$ and $\boldsymbol{\theta}$ .Here, $k = 30$ , $n = 15$. Each column of  ${{\bf{P}}_{{\rm{acc}}}}$ records the classification accuracy of an SVM model on 16 tasks associated with the same color point column in Figure (a). Each row of ${{\bf{P}}_{{\rm{acc}}}}$ records the classification accuracy of 30 SVM models on a single task. The row values are sequentially concatenated to obtain the task curve shown in Figure (b).}\label{fig11}
\end{figure}

Based on the assumptions stated above, we choose the SVM (Linear) model on the Iris dataset to perform the anti-noise experiment depicted in Figure \ref{fig9}. $k$ SVMs with relative ability values $\boldsymbol{\theta}$  were obtained by adjusting the parameters $C$   (Table \ref{tab3}). $\boldsymbol{\theta} \in {\mathbb{R}^k}$  take $k$   values evenly from 0 to 1. Each SVM tests on $n$  noisy tasks, and get ${\bf{P}}_{{\rm{acc}}}^j \in {\mathbb{R}^n}(j = 1,2,...,k)$ .  $k$-group ${\bf{P}}_{{\rm{acc}}}^j$  is sorted from small to large according to its largest element, and we have ${{\bf{P}}_{{\rm{acc}}}} = [{\bf{P}}_{{\rm{acc}}}^1,{\bf{P}}_{{\rm{acc}}}^2,...,{\bf{P}}_{{\rm{acc}}}^k] \in {\mathbb{R}^{n \times k}}$. Figure \ref{fig11} shows the mapping of ${{\bf{P}}_{{\rm{acc}}}}$  and $\boldsymbol{\theta}$ .

Observe that the shape of the curve in Figure \ref{fig11}(b) is more consistent with that of the Sigmoid function, but the upper and lower bounds of the task curve are variable; thus, equation (\ref{eq29}) is adopted as the fitting function.

\begin{equation}
	{P_{acc}} = \frac{{u - l}}{{1 + \exp ( - a*(\theta  - b))}} + u\label{eq29}
\end{equation}

\begin{figure}[h]%
	\centering
	\includegraphics[width=0.9\textwidth]{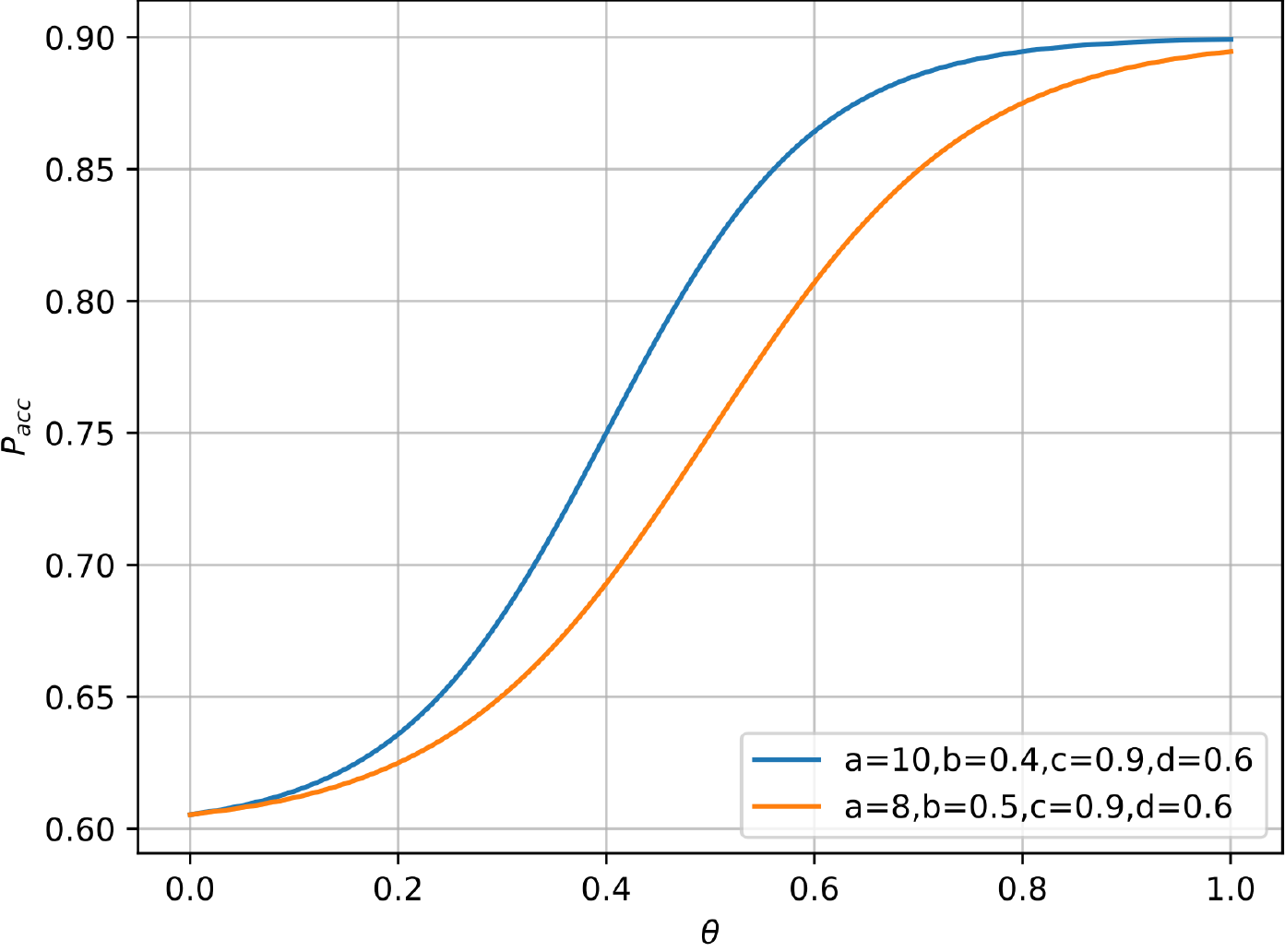}
	\caption{Fitting function plot.}\label{fig12}
\end{figure}

Plotting the curve of equation (\ref{eq29}) in Figure \ref{fig12}, we explore the properties of this function.

The parameters $u$ , $l$ , $a$ , and $b$ reflect the function properties as follows.

1) $u$ and $l$  can represent the classifier's upper and lower bounds of recognition accuracy on a set of recognition tasks, respectively.

2) $(u - l)*a$ reflects the slope of the function. The flatter the function, the harder the task and the lower the recognition accuracy.

3) $b$ affects the right shift rate of the function. The larger the right shift magnitude, the more difficult the task is, and the less accurate the classifier is.

\begin{figure}[h]%
	\centering
	\includegraphics[width=0.9\textwidth]{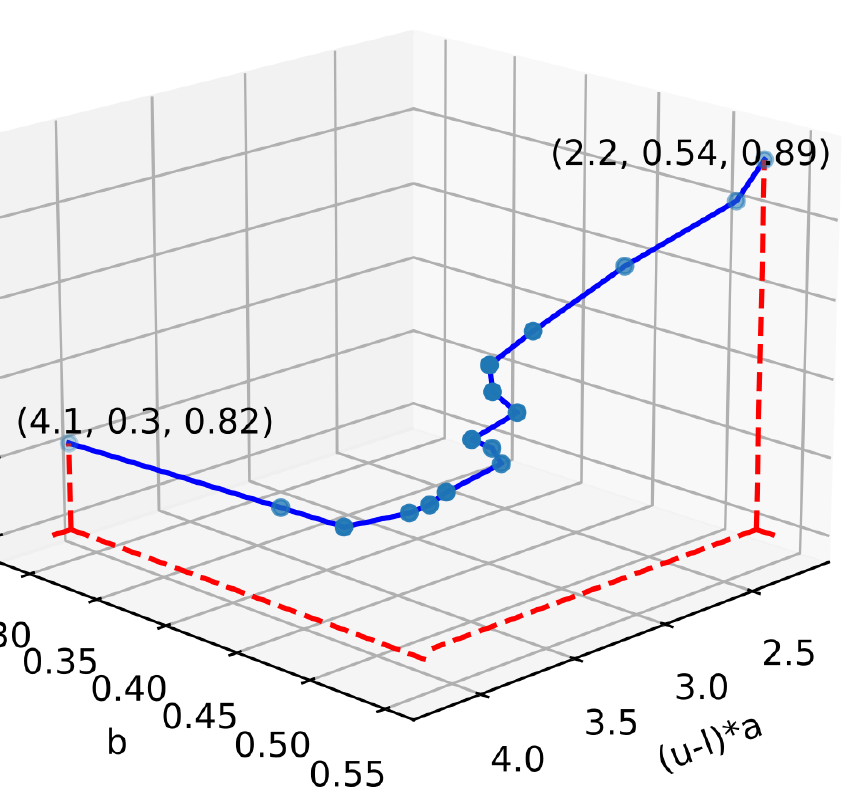}
	\caption{ the relationship between ${R_{\rm{S}}}$ , $u$ ,  $l$ , $a$ , and $b$ on the Iris dataset. The lowest  ${R_{\rm{S}}}$ value corresponds to the lowest $b$  and the steepest slope ($(u - l)*a$ ), indicating that the task is simple. The highest ${R_{\rm{S}}}$  value corresponds to the highest $b$  and the flattest slope ( $(u - l)*a$ ), indicating that the task is the most challenging.}\label{fig13}
\end{figure}

Then we research the relationship between ${R_{\rm{S}}}$ , $u$ ,  $l$ , $a$ , and $b$ , where  $u$ and $l$  are determined by the upper and lower limits of classifier performance; while $a$  and  $b$ depend on the difficulty of the task. The relationship verified on the Iris dataset is given by Figure \ref{fig13}.

Then we employ the polynomial fitting approach, with $a$  and $b$   represented by ${R_{\rm{S}}}$ 

\begin{align}
	{f_a}({R_{\rm{S}}}) = {h_0} + {h_1}{R_{\rm{S}}} + {h_2}R_{\rm{S}}^2 \nonumber \\
	{f_b}({R_{\rm{S}}}) = {p_0} + {p_1}{R_{\rm{S}}} + {p_2}R_{\rm{S}}^{\rm{2}}.\label{eq30}
\end{align}

The mapping function ${f^{ - 1}}( \cdot )$  from classification accuracy to classifier ability with separability  ${R_{\rm{S}}}$  as a parameter is now obtained.

\begin{equation}
{P_{acc}} = {f^{ - 1}}(\theta ;{R_S}) = \frac{{u - l}}{{1 + \exp ( - {f_a}({R_S})*(\theta  - {f_b}({R_S})))}} + u\label{eq31}
\end{equation}

To examine the validity of this mapping function, we need to substitute the recognition accuracy of another classifier into the equation (\ref{eq31}) to ensure the uniqueness of its recognition ability value, demonstrating that the recognition ability value exists as an inherent property of the classifier.

\begin{figure}[h]%
	\centering
	\includegraphics[width=0.9\textwidth]{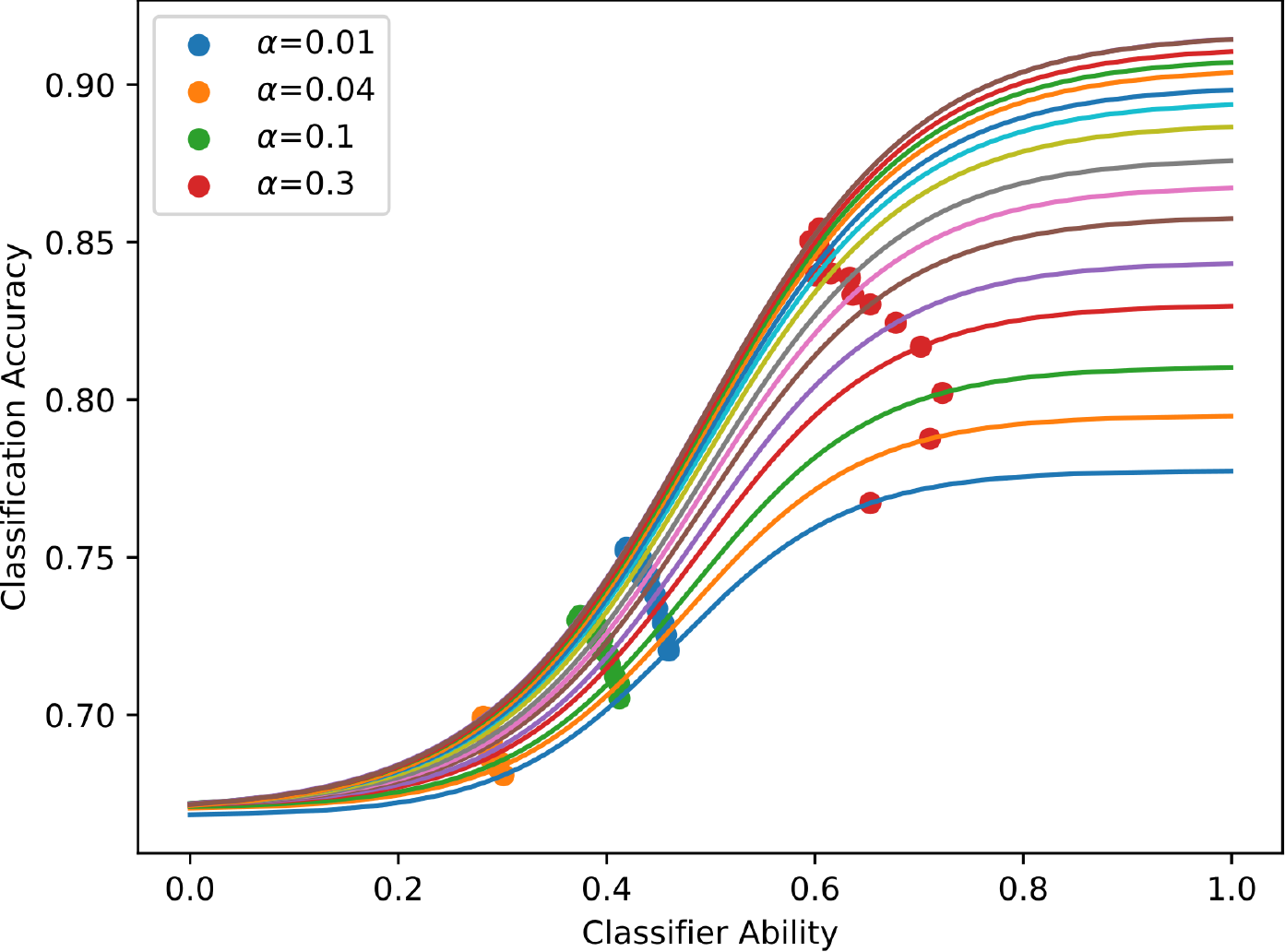}
	\caption{The fitting task curves and the mapping points. }\label{fig14}
\end{figure}

Figure \ref{fig14} shows the results of fitting the task curve with the SVM model as a reference and an evaluation of the LR model’s recognition ability on this curve.

This evaluation method has a high assessment accuracy in the middle of the task curve. The evaluation of the LR model ability values for $\alpha=0.01$  and $\alpha=0.1$ are distributed over a small interval, and a set of recognition accuracies essentially map to a unique recognition ability value. Whereas at the two ends of the curve, a slight change in recognition accuracy may bring about a significant deviation in recognition ability due to the presence of the saturation zone.

\subsection{CNN layers’ performance evaluated by data separability}\label{sec4.3}

As an extension of machine learning classifiers, deep learning classifiers have greatly improved recognition performance but are not satisfactory in model interpretability. The abstraction lies in the fact that we cannot fully grasp the substantive role played by each module in the deep network. The convolution module, for example, is widely believed to play the role of feature extraction, enabling the final output data to be more separable. Nonetheless, is it better to have more convolutional modules? And is the contribution of each convolutional module to the extraction of potential features consistent? How do we measure this contribution? To answer the above questions, in this section, we design modular classifier recognition performance evaluation methods to evaluate the performance of each convolutional component of the CNN.

Effective separability indices are invaluable for the performance evaluation of radar signal classification algorithms \cite{bib28}. Compared to the classification of optical images, radar target recognition requires more of a metric to clearly understand and grasp the substantial effect of each module when using a deep neural network. This is because the radar images are not very understandable for the non-expert. And it is more difficult to identify the classes to which they belong after their semantic features are extracted by the convolutional layer, which makes it more difficult to judge the meaning of the existence of each module. The evaluation method proposed is to insert a feature separability analysis module after each convolutional block to monitor its performance, and the separability measure used is RS.

We use the typical SAR image MSTAR\footnote{The URL for downloading the dataset:\url{https://www.sdms.afrl.af.mil/datasets/mstar/}} as the experimental data, where the SOC dataset is shown in Table \ref{tab5}.

\begin{table}[h]
	\begin{center}
		\begin{minipage}{\textwidth}
			\caption{Example of a lengthy table which is set to full textwidth}\label{tab5}
			\begin{tabular*}{\textwidth}{@{\extracolsep{\fill}}lccccc@{\extracolsep{\fill}}}
				\toprule%
				 \multicolumn{2}{@{}c@{}}{}&\multicolumn{2}{@{}c@{}}{Train} & \multicolumn{2}{@{}c@{}}{test}\\
				  \cmidrule{3-4}\cmidrule{5-6}%
				Class & Serial No. & Depression & No. Images & Depression & No. Images  \\
				\midrule
				BMP-2 & 9563 & 17° & 233 & 15° & 196 \\
				BTR-70 & C71 & 17° & 233 & 15° & 196 \\
				T-72 & 132 & 17° & 232 & 15° & 196 \\
				BTR-60 & k10yt7532 & 17° & 256 & 15° & 195 \\
				2S1 & b01 & 17° & 299 & 15° & 274 \\
				BRDM-2 & E-71 & 17° & 298 & 15° & 274 \\
				D7 & 92v13015 & 17° & 299 & 15° & 274 \\
				T-62 & A51 & 17° & 299 & 15° & 273 \\
				ZIL-131 & E12 & 17° & 299 & 15° & 274 \\
				ZSU-234 & d08 & 17° & 299 & 15° & 274 \\
				\botrule
			\end{tabular*}
		\end{minipage}
	\end{center}
\end{table}

\begin{figure}[H]%
	\centering
	\includegraphics[width=0.9\textwidth]{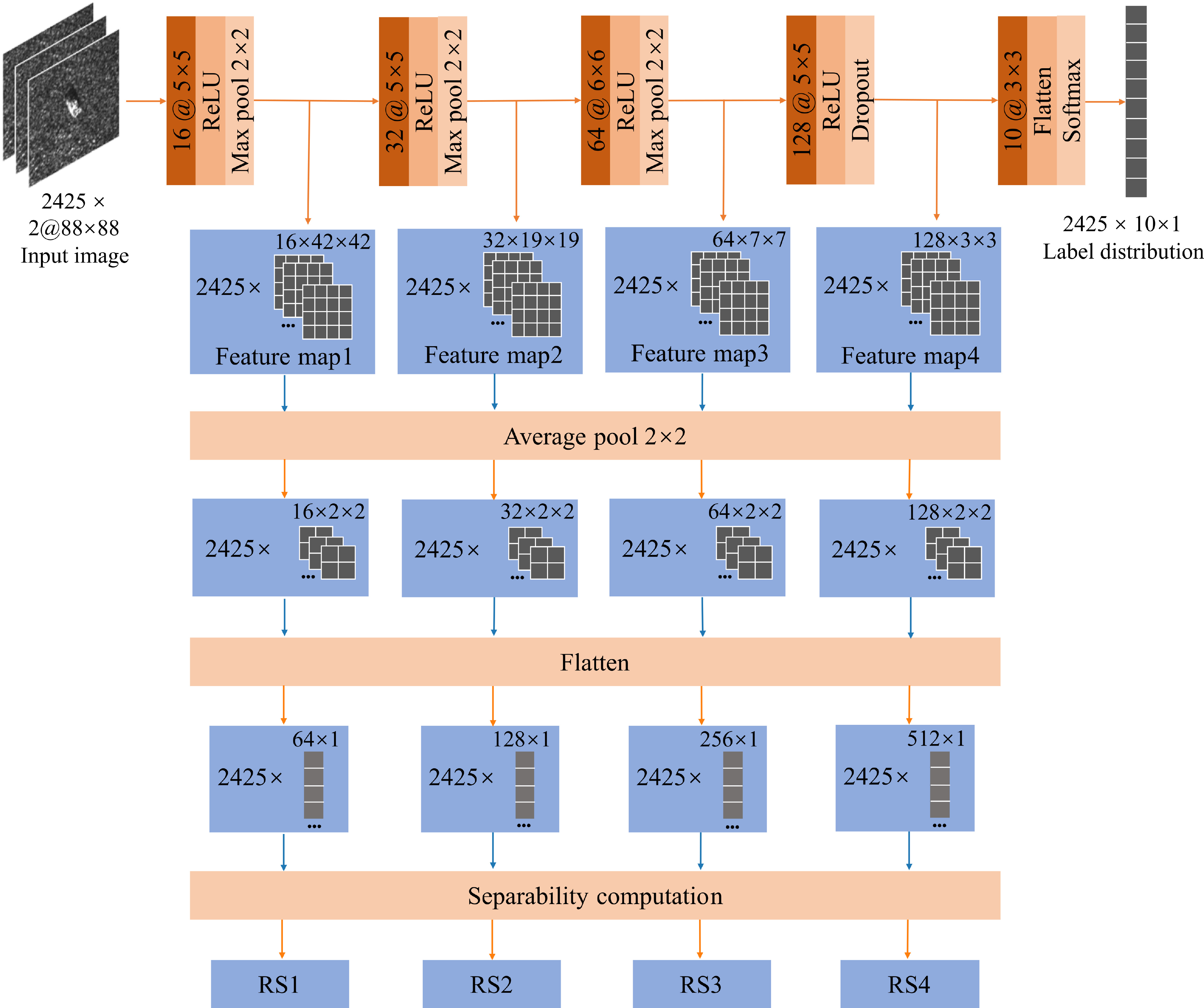}
	\caption{Network architecture and feature separability analysis module. }\label{fig15}
\end{figure}

\begin{figure}[H]%
	\centering
	\includegraphics[width=0.9\textwidth]{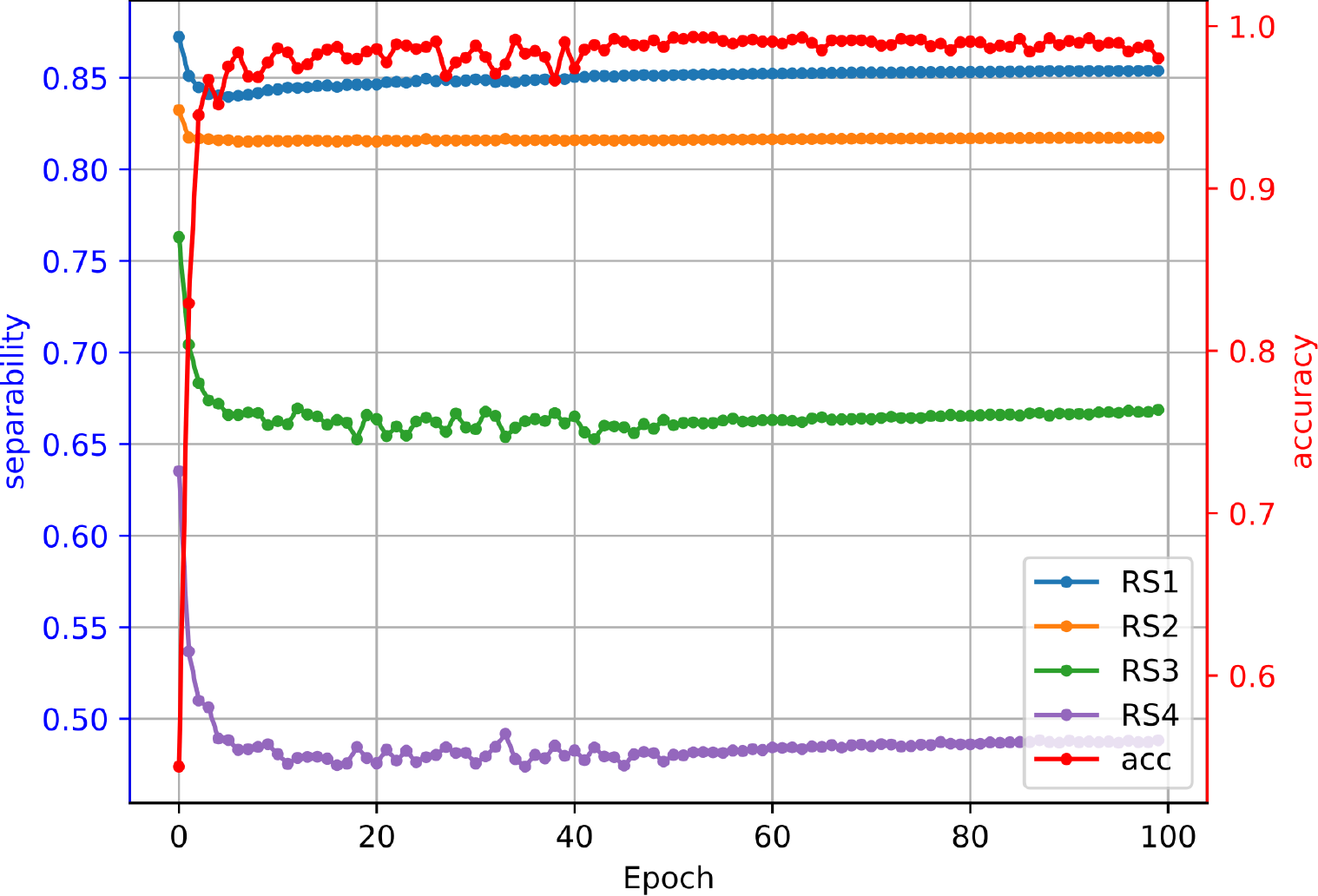}
	\caption{Classification accuracy and feature separability analysis results. }\label{fig16}
\end{figure}
On the dataset presented in Table \ref{tab5}, we evaluate the performance of some convolutional blocks of CNN provided by Chen et al. \cite{bib29}. And to reduce the computational effort of RS, we apply a 2*2 average pooling to the feature map. The network structure and the feature separability analysis module are shown in Figure \ref{fig15}.

After 100 epochs of training, the convergence of recognition accuracy on the test set and the variation of feature separability extracted by each convolutional block are set out in Figure \ref{fig16}.

The most striking result from Figure \ref{fig16} is that the separability of features extracted by each convolutional block keeps step with the final classification accuracy. When classification accuracy improves dramatically, the RS value falls precipitously. And when the classification accuracy converges, the RS value becomes steady. Furthermore, we also find that the convolutional block in a deeper network has a more significant function in extracting a more separable feature with a lower RS value (${\rm{RS1 > RS2 > RS3 > RS4}}$). And the deeper feature map exhibits a wider dynamic range of RS value ($\Delta {\rm{RS1}} < \Delta {\rm{RS2}} < \Delta {\rm{RS3}} < \Delta {\rm{RS4}}$). The evidence above claims that all four convolutional blocks perform well in this network, and the deeper block does better work.

\section{Conclusion}\label{sec5}

We propose a new separability measure based on the data encoding rate. The measure can capture the relative relationship between the expansion rate of the sample distribution in the whole space and the subspace's compression rate. We validate the effectiveness of the proposed measure on a typical synthetic two-class dataset and confirm its positive correlation with the classification accuracy in a series of noisy tasks constructed from real datasets. We designed machine learning and deep learning classifier model evaluation methods based on the above two basic argumentation experiments. We build a mapping model for machine learning classifiers from classification accuracy to classifier ability. In the model, the task difficulty is characterized by the measure, and the classification accuracy assesses the value of the classifier capability as its inherent properties with a separability measure as a parameter. The method applies to a subset of classifiers under certain assumptions. For deep learning classifiers, we use a modular evaluation approach. The ability of each convolutional block at different depths of the network to extract separable features is discussed using the proposed measure.

Data separability quantification provides some basis for analyzing, understanding, and enhancing model performance. This paper focuses on its application in the recognition field. In fact, it can also be applied to evaluate clustering results \cite{bib30}, understand the demerit of each feature \cite{bib31,bib32}, provide a theory for building multi-classifier decision \cite{bib33}, reduce data complexity as a loss function \cite{bib34}, and other fields. Furthermore, a more comprehensive metric should be served as a general standard. The values are still comparable when comparing datasets with different sample sizes and feature dimensions. The existing separability measures still lack the above generalization properties, and this is a point that deserves further study.

\section*{Declarations}

\begin{itemize}
\item \textbf{Conflict of interests} The authors declare that they have no conflict of interest.
\item \textbf{Funding} This work was supported in part by National Natural Science Foundation of China under Grants 62022091(Corresponding author:
Xinyu Zhang.)
\item \textbf{Data Availability Statements} The data in \ref{sec4.1} and \ref{sec4.2} that support the findings of this study are openly available in UCI Machine Learning Repository at \url{https://archive.ics.uci.edu}, reference number \cite{bib28}. And the network we analyzed in \ref{sec4.3} is derived at \url{https://github.com/fudanxu/MSTAR-AConvNet}, reference number \cite{bib29}.
\end{itemize}

\bibliography{sn-bibliography}


\begin{thebibliography}{34}
\ifx \bisbn   \undefined \def \bisbn  #1{ISBN #1}\fi
\ifx \binits  \undefined \def \binits#1{#1}\fi
\ifx \bauthor  \undefined \def \bauthor#1{#1}\fi
\ifx \batitle  \undefined \def \batitle#1{#1}\fi
\ifx \bjtitle  \undefined \def \bjtitle#1{#1}\fi
\ifx \bvolume  \undefined \def \bvolume#1{\textbf{#1}}\fi
\ifx \byear  \undefined \def \byear#1{#1}\fi
\ifx \bissue  \undefined \def \bissue#1{#1}\fi
\ifx \bfpage  \undefined \def \bfpage#1{#1}\fi
\ifx \blpage  \undefined \def \blpage #1{#1}\fi
\ifx \burl  \undefined \def \burl#1{\textsf{#1}}\fi
\ifx \doiurl  \undefined \def \doiurl#1{\url{https://doi.org/#1}}\fi
\ifx \betal  \undefined \def \betal{\textit{et al.}}\fi
\ifx \binstitute  \undefined \def \binstitute#1{#1}\fi
\ifx \binstitutionaled  \undefined \def \binstitutionaled#1{#1}\fi
\ifx \bctitle  \undefined \def \bctitle#1{#1}\fi
\ifx \beditor  \undefined \def \beditor#1{#1}\fi
\ifx \bpublisher  \undefined \def \bpublisher#1{#1}\fi
\ifx \bbtitle  \undefined \def \bbtitle#1{#1}\fi
\ifx \bedition  \undefined \def \bedition#1{#1}\fi
\ifx \bseriesno  \undefined \def \bseriesno#1{#1}\fi
\ifx \blocation  \undefined \def \blocation#1{#1}\fi
\ifx \bsertitle  \undefined \def \bsertitle#1{#1}\fi
\ifx \bsnm \undefined \def \bsnm#1{#1}\fi
\ifx \bsuffix \undefined \def \bsuffix#1{#1}\fi
\ifx \bparticle \undefined \def \bparticle#1{#1}\fi
\ifx \barticle \undefined \def \barticle#1{#1}\fi
\bibcommenthead
\ifx \bconfdate \undefined \def \bconfdate #1{#1}\fi
\ifx \botherref \undefined \def \botherref #1{#1}\fi
\ifx \url \undefined \def \url#1{\textsf{#1}}\fi
\ifx \bchapter \undefined \def \bchapter#1{#1}\fi
\ifx \bbook \undefined \def \bbook#1{#1}\fi
\ifx \bcomment \undefined \def \bcomment#1{#1}\fi
\ifx \oauthor \undefined \def \oauthor#1{#1}\fi
\ifx \citeauthoryear \undefined \def \citeauthoryear#1{#1}\fi
\ifx \endbibitem  \undefined \def \endbibitem {}\fi
\ifx \bconflocation  \undefined \def \bconflocation#1{#1}\fi
\ifx \arxivurl  \undefined \def \arxivurl#1{\textsf{#1}}\fi
\csname PreBibitemsHook\endcsname

\bibitem{bib1}
\begin{barticle}
\bauthor{\bsnm{Cano}, \binits{J.R.}}:
\batitle{Analysis of data complexity measures for classification}.
\bjtitle{Expert Systems With Applications}
\bvolume{40}(\bissue{12}),
\bfpage{4820}--\blpage{4831}
(\byear{2013}).
\bcomment{\url{https://doi.org/10.1016/j.eswa.2013.02.025}}
\end{barticle}
\endbibitem

\bibitem{bib2}
\begin{barticle}
\bauthor{\bsnm{Wicaksonoa}, \binits{P.}},
\bauthor{\bsnm{Aryagunab}, \binits{P.A.}}:
\batitle{Analyses of inter-class spectral separability and classification
  accuracy of benthic habitat mapping using multispectral image}.
\bjtitle{Remote Sensing Applications: Society and Environment}
\bvolume{19},
\bfpage{100335}
(\byear{2020}).
\bcomment{\url{https://doi.org/10.1016/j.rsase.2020.100335}}
\end{barticle}
\endbibitem

\bibitem{bib3}
\begin{barticle}
\bauthor{\bsnm{Li}, \binits{S.}},
\bauthor{\bsnm{Hao}, \binits{Q.}},
\bauthor{\bsnm{Gao}, \binits{G.}},
\bauthor{\bsnm{Kang}, \binits{X.}}:
\batitle{The effect of ground truth on performance evaluation of hyperspectral
  image classification}.
\bjtitle{IEEE Transactions on Geoscience and Remote Sensing}
\bvolume{56}(\bissue{12}),
\bfpage{7195}--\blpage{7206}
(\byear{2018}).
\bcomment{\url{https://doi.org/10.1109/TGRS.2018.2849225}}
\end{barticle}
\endbibitem

\bibitem{bib4}
\begin{barticle}
\bauthor{\bsnm{Oprea}, \binits{M.}}:
\batitle{A general framework and guidelines for benchmarking computational
  intelligence algorithms applied to forecasting problems derived from an
  application domain-oriented survey}.
\bjtitle{Applied Soft Computing}
\bvolume{89},
\bfpage{106103}
(\byear{2020}).
\bcomment{\url{https://doi.org/10.1016/j.asoc.2020.106103}}
\end{barticle}
\endbibitem

\bibitem{bib5}
\begin{botherref}
\oauthor{\bsnm{Lee}, \binits{Y.}},
\oauthor{\bsnm{Lee}, \binits{J.}},
\oauthor{\bsnm{Hwang}, \binits{S.J.}},
\oauthor{\bsnm{Yang}, \binits{E.}},
\oauthor{\bsnm{Choi}, \binits{S.}}:
Neural complexity measures.
In: Proceedings of the 34th International Conference on Neural Information
  Processing Systems,9713--9724
(2020)
\end{botherref}
\endbibitem

\bibitem{bib6}
\begin{barticle}
\bauthor{\bsnm{Zhang}, \binits{C.}},
\bauthor{\bsnm{Samy}, \binits{B.}},
\bauthor{\bsnm{Moritz}, \binits{H.}},
\bauthor{\bsnm{Benjamin}, \binits{R.}},
\bauthor{\bsnm{Oriol}, \binits{V.}}:
\batitle{Understanding deep learning (still) requires rethinking
  generalization}.
\bjtitle{Communications of the ACM}
\bvolume{64}(\bissue{3}),
\bfpage{107}--\blpage{115}
(\byear{2021}).
\bcomment{\url{https://doi.org/10.1145/3446776}}
\end{barticle}
\endbibitem

\bibitem{bib7}
\begin{barticle}
\bauthor{\bsnm{Hossin}, \binits{M.}},
\bauthor{\bsnm{Sulaiman}, \binits{M.N.}}:
\batitle{A review on evaluation metrics for data classification evaluations}.
\bjtitle{International Journal of Data Mining and Knowledge Management Process}
\bvolume{5}(\bissue{2}),
\bfpage{1}--\blpage{11}
(\byear{2015}).
\bcomment{\url{https://doi.org/10.5121/ijdkp.2015.5201}}
\end{barticle}
\endbibitem

\bibitem{bib8}
\begin{barticle}
\bauthor{\bsnm{Yu}, \binits{S.}},
\bauthor{\bsnm{Li}, \binits{X.}},
\bauthor{\bsnm{Feng}, \binits{Y.}},
\bauthor{\bsnm{Zhang}, \binits{X.}},
\bauthor{\bsnm{Chen}, \binits{S.}}:
\batitle{An instance-oriented performance measure for classification}.
\bjtitle{Information Sciences}
\bvolume{580},
\bfpage{598}--\blpage{619}
(\byear{2021}).
\bcomment{\url{https://doi.org/10.1016/j.ins.2021.08.094}}
\end{barticle}
\endbibitem

\bibitem{bib9}
\begin{bchapter}
\bauthor{\bsnm{Fernández}, \binits{A.}},
\bauthor{\bsnm{García}, \binits{M.} \bsuffix{S.and~Galar}},
\bauthor{\bsnm{Prati}, \binits{R.C.}},
\bauthor{\bsnm{Krawczyk}, \binits{B.}},
\bauthor{\bsnm{Herrera}, \binits{F.}}:
\bctitle{Data intrinsic characteristics}.
In: \bbtitle{Learning from Imbalanced Data Sets},
pp. \bfpage{253}--\blpage{277}.
\bpublisher{Springer},
\blocation{Cham}
(\byear{2018}).
\bcomment{\url{https://doi.org/10.1007/978-3-319-98074-4_10}}
\end{bchapter}
\endbibitem

\bibitem{bib10}
\begin{barticle}
\bauthor{\bsnm{Bello1}, \binits{M.}},
\bauthor{\bsnm{Nápoles}, \binits{G.}},
\bauthor{\bsnm{Vanhoof}, \binits{K.}},
\bauthor{\bsnm{Bello}, \binits{R.}}:
\batitle{Data quality measures based on granular computing for multi-label
  classification}.
\bjtitle{Information Sciences}
\bvolume{560},
\bfpage{51}--\blpage{67}
(\byear{2021}).
\bcomment{\url{https://doi.org/10.1016/j.ins.2021.01.027}}
\end{barticle}
\endbibitem

\bibitem{bib11}
\begin{botherref}
\oauthor{\bsnm{Li}, \binits{C.}},
\oauthor{\bsnm{Wang}, \binits{B.}}:
Fisher Linear Discriminant Analysis.
CCIS Northeastern University
(2014)
\end{botherref}
\endbibitem

\bibitem{bib12}
\begin{barticle}
\bauthor{\bsnm{Hossin}, \binits{M.}},
\bauthor{\bsnm{Sulaiman}, \binits{M.N.}}:
\batitle{A framework for dynamic classifier selection oriented by the
  classification problem difficulty}.
\bjtitle{Pattern Recognition}
\bvolume{76}(\bissue{1}),
\bfpage{175}--\blpage{190}
(\byear{2018}).
\bcomment{\url{https://doi.org/10.1016/j.patcog.2017.10.038}}
\end{barticle}
\endbibitem

\bibitem{bib13}
\begin{barticle}
\bauthor{\bsnm{Ho}, \binits{T.K.}},
\bauthor{\bsnm{Basu}, \binits{M.}}:
\batitle{Complexity measures of supervised classification problems}.
\bjtitle{IEEE Transactions on Pattern Analysis and Machine Intelligence}
\bvolume{24}(\bissue{3}),
\bfpage{289}--\blpage{300}
(\byear{2002}).
\bcomment{\url{https://doi.org/10.1109/34.990132}}
\end{barticle}
\endbibitem

\bibitem{bib14}
\begin{barticle}
\bauthor{\bsnm{Lorena}, \binits{A.C.}},
\bauthor{\bsnm{Garcia}, \binits{L.P.F.}},
\bauthor{\bsnm{Lehmann}, \binits{J.}},
\bauthor{\bsnm{Souto}, \binits{M.C.P.}},
\bauthor{\bsnm{Ho}, \binits{T.K.}}:
\batitle{How complex is your classification problem?: A survey on measuring
  classification complexity}.
\bjtitle{ACM Computing Surveys}
\bvolume{52}(\bissue{5}),
\bfpage{1}--\blpage{34}
(\byear{2019}).
\bcomment{\url{https://doi.org/10.1145/3347711}}
\end{barticle}
\endbibitem

\bibitem{bib15}
\begin{barticle}
\bauthor{\bsnm{Ferraro}, \binits{M.B.}},
\bauthor{\bsnm{Giordani}, \binits{P.}}:
\batitle{A review and proposal of (fuzzy) clustering for nonlinearly separable
  data}.
\bjtitle{International Journal of Approximate Reasoning}
\bvolume{115}(\bissue{C}),
\bfpage{13}--\blpage{31}
(\byear{2019}).
\bcomment{\url{https://doi.org/10.1016/j.ijar.2019.09.004}}
\end{barticle}
\endbibitem

\bibitem{bib16}
\begin{botherref}
\oauthor{\bsnm{Aggarwal}, \binits{C.C.}},
\oauthor{\bsnm{Hinneburg}, \binits{A.}},
\oauthor{\bsnm{Keim}, \binits{D.A.}}:
On the Surprising Behavior of Distance Metrics in High Dimensional Space.
In: Van den Bussche, J., Vianu, V. (eds) Database Theory — ICDT 2001. ICDT
  2001. Lecture Notes in Computer Science, vol 1973. Springer, Berlin,
  Heidelberg. \url{https://doi.org/10.1007/3-540-44503-X_27}
(2001)
\end{botherref}
\endbibitem

\bibitem{bib17}
\begin{bbook}
\bauthor{\bsnm{Cover}, \binits{T.M.}},
\bauthor{\bsnm{Thomas}, \binits{J.A.}}:
\bbtitle{Elements of Information Theory}.
\bpublisher{Wiley Interscience},
\blocation{USA}
(\byear{2006})
\end{bbook}
\endbibitem

\bibitem{bib18}
\begin{botherref}
\oauthor{\bsnm{Madiman}, \binits{M.}},
\oauthor{\bsnm{Harrison}, \binits{M.}},
\oauthor{\bsnm{Kontoyiannis}, \binits{I.}}:
Minimum description length versus maximum likelihood in lossy data compression.
In: International Symposium onInformation Theory, 2004. IEEE, Chicago.
  \url{https://doi.org/10.1109/ISIT.2004.1365499}
(2004)
\end{botherref}
\endbibitem

\bibitem{bib19}
\begin{barticle}
\bauthor{\bsnm{Ma}, \binits{Y.}},
\bauthor{\bsnm{Derksen}, \binits{H.}},
\bauthor{\bsnm{Hong}, \binits{W.}}:
\batitle{Segmentation of multivariate mixed data via lossy data coding and
  compression}.
\bjtitle{IEEE Transactions on Pattern Analysis and Machine Intelligence}
\bvolume{29}(\bissue{9}),
\bfpage{1546}--\blpage{1562}
(\byear{2007}).
\bcomment{\url{https://doi.org/10.1109/TPAMI.2007.1085}}
\end{barticle}
\endbibitem

\bibitem{bib20}
\begin{bchapter}
\bauthor{\bsnm{Guan}, \binits{S.}},
\bauthor{\bsnm{Loew}, \binits{M.}}:
\bctitle{A novel intrinsic measure of data separability}.
In: \bbtitle{Applied Intelligence},
(\byear{2022}).
\bcomment{\url{https://doi.org/10.1007/s10489-022-03395-6}}
\end{bchapter}
\endbibitem

\bibitem{bib21}
\begin{barticle}
\bauthor{\bsnm{Shannon}, \binits{C.E.}}:
\batitle{A mathematical theory of communication}.
\bjtitle{The Bell System Technical Journal}
\bvolume{27}(\bissue{3}),
\bfpage{379}--\blpage{423}
(\byear{1948}).
\bcomment{\url{https://doi.org/10.1002/j.1538-7305.1948.tb01338.x}}
\end{barticle}
\endbibitem

\bibitem{bib22}
\begin{botherref}
\oauthor{\bsnm{Macdonald}, \binits{J.}},
\oauthor{\bsnm{Wäldchen}, \binits{S.}},
\oauthor{\bsnm{Hauch}, \binits{S.}},
\oauthor{\bsnm{Kutyniok}, \binits{G.}}:
A rate-distortion framework for explaining neural network decisions.
Statistics
(2019).
\url{https://doi.org/10.48550/arXiv.1905.11092}
\end{botherref}
\endbibitem

\bibitem{bib23}
\begin{botherref}
\oauthor{\bsnm{Wu}, \binits{Z.}},
\oauthor{\bsnm{Baek}, \binits{C.}},
\oauthor{\bsnm{You}, \binits{C.}},
\oauthor{\bsnm{Ma}, \binits{Y.}}:
Incremental learning via rate reduction.
In: 2021 IEEE/CVF Conference on Computer Vision and Pattern Recognition (CVPR).
  IEEE, Nashville. \url{https://doi.org/10.1109/CVPR46437.2021.00118}
(2021)
\end{botherref}
\endbibitem

\bibitem{bib24}
\begin{botherref}
\oauthor{\bsnm{Elizondo}, \binits{D.A.}},
\oauthor{\bsnm{Birkenhead}, \binits{R.}},
\oauthor{\bsnm{Gamez}, \binits{M.}},
\oauthor{\bsnm{Garcia}, \binits{N.}},
\oauthor{\bsnm{Alfaro}, \binits{E.}}:
Linear separability and classification complexity
\textbf{39}(9),
7796--7807
(2012).
\url{https://doi.org/10.1016/j.eswa.2012.01.090}
\end{botherref}
\endbibitem

\bibitem{bib25}
\begin{botherref}
\oauthor{\bsnm{Yu}, \binits{Y.}},
\oauthor{\bsnm{Chan}, \binits{K.H.R.}},
\oauthor{\bsnm{You}, \binits{C.}},
\oauthor{\bsnm{Song}, \binits{C.}},
\oauthor{\bsnm{Ma}, \binits{Y.}}:
Learning diverse and discriminative representations via the principle of
  maximal coding rate reduction
(2020).
\url{https://doi.org/10.48550/arXiv.2006.08558}
\end{botherref}
\endbibitem

\bibitem{bib26}
\begin{barticle}
\bauthor{\bsnm{Luor}, \binits{D.C.}}:
\batitle{A comparative assessment of data standardization on support vector
  machine for classification problems}.
\bjtitle{Intelligent Data Analysis}
\bvolume{19}(\bissue{3}),
\bfpage{529}--\blpage{546}
(\byear{2015}).
\bcomment{\url{https://doi.org/10.3233/IDA-150730}}
\end{barticle}
\endbibitem

\bibitem{bib27}
\begin{barticle}
\bauthor{\bsnm{Mishra}, \binits{A.K.}}:
\batitle{Separability indices and their use in radar signal based target
  recognition}.
\bjtitle{IEICE Electronics Express}
\bvolume{6}(\bissue{14}),
\bfpage{1000}--\blpage{1005}
(\byear{2009}).
\bcomment{\url{https://doi.org/10.1587/elex.6.1000}}
\end{barticle}
\endbibitem

\bibitem{bib28}
\begin{botherref}
\oauthor{\bsnm{Lichman}, \binits{M.e.a.}}:
Uci machine learning reposit.
\url{https://archive.ics.uci.edu/ml/datasets.php}
(2013)
\end{botherref}
\endbibitem

\bibitem{bib29}
\begin{barticle}
\bauthor{\bsnm{Sizhe}, \binits{C.}},
\bauthor{\bsnm{Haipeng}, \binits{W.}},
\bauthor{\bsnm{Feng}, \binits{X.}},
\bauthor{\bsnm{Yaqiu}, \binits{J.}}:
\batitle{Target classification using the deep convolutional networks for sar
  images}.
\bjtitle{IEEE Transactions on Geoscience and Remote Sensing}
\bvolume{54}(\bissue{8}),
\bfpage{4806}--\blpage{4817}
(\byear{2016}).
\bcomment{\url{https://doi.org/10.1109/TGRS.2016.2551720}}
\end{barticle}
\endbibitem

\bibitem{bib30}
\begin{barticle}
\bauthor{\bsnm{Arbelaitz}, \binits{O.}},
\bauthor{\bsnm{Gurrutxaga}, \binits{I.}},
\bauthor{\bsnm{Muguerza}, \binits{J.}},
\bauthor{\bsnm{M.Pérez}, \binits{J.}},
\bauthor{\bsnm{Perona}, \binits{I.}}:
\batitle{An extensive comparative study of cluster validity indices}.
\bjtitle{Pattern Recognition}
\bvolume{46}(\bissue{1}),
\bfpage{243}--\blpage{256}
(\byear{2013}).
\bcomment{\url{https://doi.org/10.1016/j.patcog.2012.07.021}}
\end{barticle}
\endbibitem

\bibitem{bib31}
\begin{botherref}
\oauthor{\bsnm{Dong}, \binits{N.T.}},
\oauthor{\bsnm{Khosla}, \binits{M.}}:
Revisiting feature selection with data complexity.
In: 2020 IEEE 20th International Conference on Bioinformatics and
  Bioengineering (BIBE). IEEE, Cincinnati, OH, USA.
  \url{https://doi.org/10.1109/BIBE50027.2020.00042}
(2020)
\end{botherref}
\endbibitem

\bibitem{bib32}
\begin{botherref}
\oauthor{\bsnm{Zhu}, \binits{Y.}},
\oauthor{\bsnm{Sun}, \binits{J.}},
\oauthor{\bsnm{Wang}, \binits{M.}},
\oauthor{\bsnm{Yao}, \binits{R.}},
\oauthor{\bsnm{Zhang}, \binits{Y.}}:
Feature separability based on the distance matrix.
In: 2017 International Conference on Orange Technologies (ICOT). IEEE,
  Singapore. \url{https://doi.org/ 10.1109/ICOT.2017.8336087}
(2017)
\end{botherref}
\endbibitem

\bibitem{bib33}
\begin{barticle}
\bauthor{\bsnm{Schilling}, \binits{A.}},
\bauthor{\bsnm{Maier}, \binits{A.}},
\bauthor{\bsnm{Gerum}, \binits{R.}},
\bauthor{\bsnm{Metzner}, \binits{C.}},
\bauthor{\bsnm{Krauss}, \binits{P.}}:
\batitle{Quantifying the separability of data classes in neural networks}.
\bjtitle{Neural networks : the official journal of the International Neural
  Network Society}
\bvolume{139},
\bfpage{278}--\blpage{293}
(\byear{2021}).
\bcomment{\url{https://doi.org/10.1016/j.neunet.2021.03.035}}
\end{barticle}
\endbibitem

\bibitem{bib34}
\begin{botherref}
\oauthor{\bsnm{Charte}, \binits{D.}},
\oauthor{\bsnm{Charte}, \binits{F.}},
\oauthor{\bsnm{Herrera}, \binits{F.}}:
Reducing data complexity using autoencoders with class-informed loss functions.
IEEE Transactions on Pattern Analysis and Machine Intelligence
(2021).
\url{https://doi.org/10.1109/TPAMI.2021.3127698}
\end{botherref}
\endbibitem

\end{thebibliography}
\end{document}